\documentclass[10pt,twocolumn,letterpaper]{article}

\usepackage[pagenumbers]{iccv} 

\usepackage{multirow}
\usepackage{booktabs}
\usepackage{makecell}

\usepackage{arydshln}
\makeatletter
\def\adl@drawiv#1#2#3{%
        \hskip.5\tabcolsep
        \xleaders#3{#2.5\@tempdimb #1{1}#2.5\@tempdimb}%
                #2\z@ plus1fil minus1fil\relax
        \hskip.5\tabcolsep}
\newcommand{\cdashlinelr}[1]{%
  \noalign{\vskip\aboverulesep
           \global\let\@dashdrawstore\adl@draw
           \global\let\adl@draw\adl@drawiv}
  \cdashline{#1}
  \noalign{\global\let\adl@draw\@dashdrawstore
           \vskip\belowrulesep}}
\makeatother

\newcommand{\dev}[2]{\makecell[c]{#1\\#2}}  

\definecolor{gold}{HTML}{FFD218}
\definecolor{silver}{HTML}{CECECE}
\definecolor{bronze}{HTML}{E5A01D}

\definecolor{iccvblue}{rgb}{0.21,0.49,0.74}
\usepackage[pagebackref,breaklinks,colorlinks,allcolors=iccvblue]{hyperref}

\def\paperID{*****} 
\def\confName{ICCV}
\def\confYear{2025}

\title{AIM 2025 Low-light RAW Video Denoising Challenge: \\Dataset, Methods and Results}

\author{
Alexander Yakovenko$^\dagger$ \quad
George Chakvetadze$^\dagger$ \quad
Ilya Khrapov$^\dagger$ \quad
Maksim Zhelezov$^\dagger$ \\
Dmitry Vatolin$^\dagger$ \quad
Radu Timofte$^\dagger$ \quad
Youngjin Oh \quad
Junhyeong Kwon \quad
Junyoung Park \\
Nam Ik Cho \quad
Senyan Xu \quad
Ruixuan Jiang \quad
Long Peng \quad
Xueyang Fu \quad
Zheng-Jun Zha \\
Xiaoping Peng \quad
Hansen Feng \quad
Zhanyi Tie \quad
Ziming Xia \quad
Lizhi Wang
}
\begin{document}
\maketitle
\let\thefootnote\relax\footnotetext{
$^\dagger$
Alexander Yakovenko, George Chakvetadze, Ilya Khrapov, Maksim Zhelezov, Dmitry Vatolin, and Radu Timofte are the challenge organizers. The other authors participated in the challenge.
Each team described its own method in the report. Appendix~\ref{sec:teams} contains the teams, affiliations, and architectures if available. \\ AIM 2025 webpage:~\url{https://cvlai.net/aim/2025}.\\
Challenge Page:~\url{https://codabench.org/competitions/8729/}.
} 

\begin{abstract}
This paper reviews the AIM~2025 (Advances in Image Manipulation) Low-Light RAW Video Denoising Challenge. The task is to develop methods that denoise low-light RAW video by exploiting temporal redundancy while operating under exposure-time limits imposed by frame rate and adapting to sensor-specific, signal-dependent noise. We introduce a new benchmark of 756 ten-frame sequences captured with 14 smartphone camera sensors across nine conditions (illumination: 1/5/10~lx; exposure: 1/24, 1/60, 1/120~s), with high-SNR references obtained via burst averaging. Participants process linear RAW sequences and output the denoised 10th frame while preserving the Bayer pattern. Submissions are evaluated on a private test set using full-reference PSNR and SSIM, with final ranking given by the mean of per-metric ranks. This report describes the dataset, challenge protocol, and submitted approaches.
\end{abstract}    
\section{Introduction}
\label{sec:intro}

Low-light RAW video denoising is critical for both consumer photography and downstream computer-vision applications. In extremely dim conditions, sensors must operate at high gain and short exposure; moreover, unlike still imaging, video imposes an upper bound on exposure time to maintain a target frame rate (e.g., 1/24-1/120\,s), preventing simple exposure increases to collect more light. As a result, read noise, photon shot noise, and sensor-specific artifacts dominate \cite{hasinoff2014photon, wei2020physics}, producing severe degradation that single-image algorithms cannot adequately address. Temporal aggregation can increase effective signal-to-noise-ratio (SNR), yet robust multi-frame denoising is challenging due to inter-frame misalignment and scene motion, and because the RAW measurements exhibit sensor-specific, signal-dependent noise characteristics.

\begin{figure}[t]
    \centering
    \includegraphics[width=\linewidth]{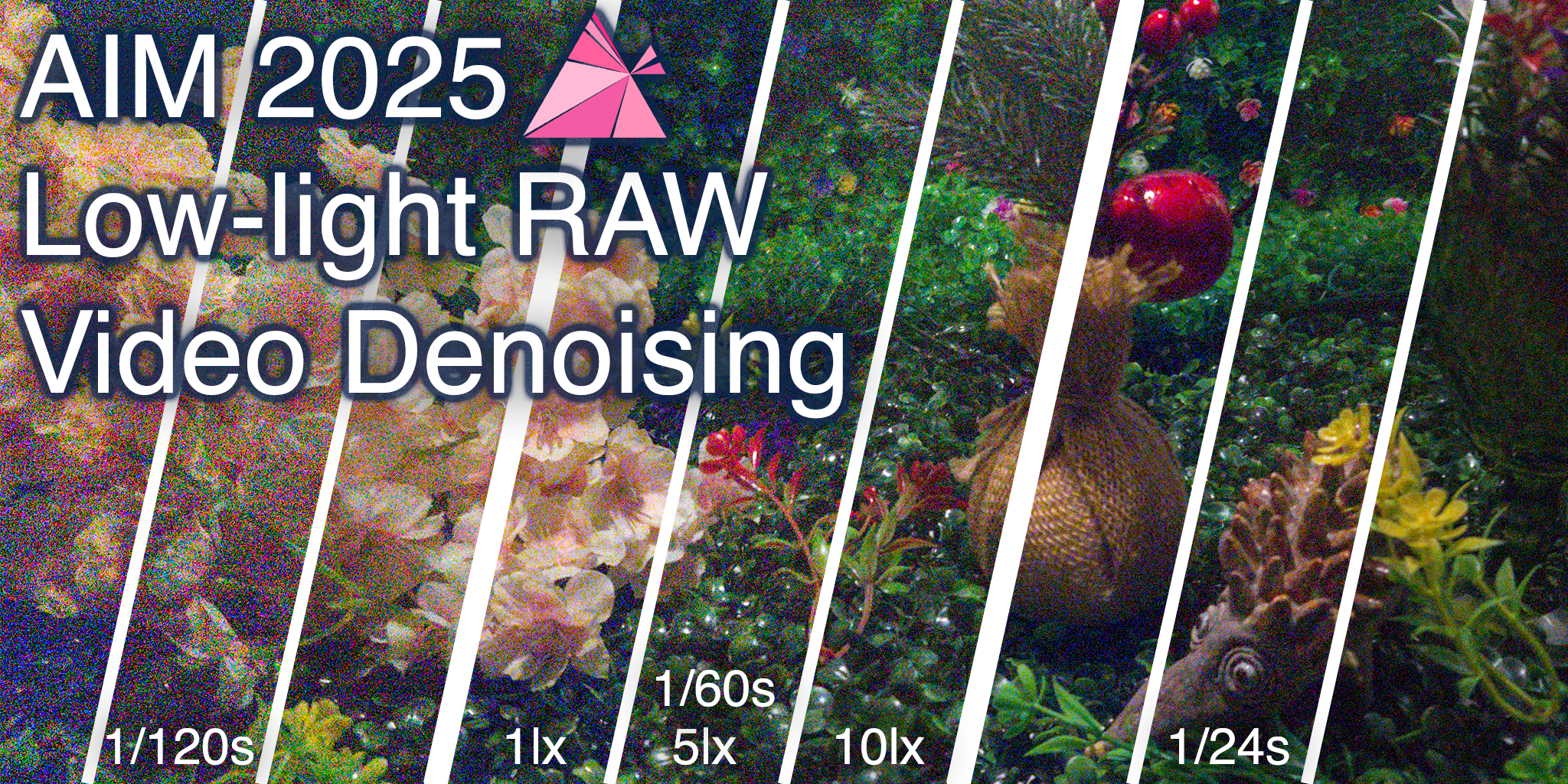}
    \caption{\textbf{Challenge teaser image.} A section of \texttt{spring} scene showcasing the different capture conditions (varying illumination and exposure). The raw \texttt{.dng} files produced by smartphone were rendered to sRGB with Adobe Photoshop \cite{adobephotoshop} for visualization with slightly improved contrast and saturation.}
    \label{fig:enter-label}
\end{figure}

Various denoising approaches have been explored. Classical non-local methods group similar patches and jointly filter them \cite{buades2005non, dabov2007image}, with temporal extensions as long-standing baselines \cite{maggioni2012video, arias2018video}. Learned models replaced hand-crafted priors with CNN/RNN designs, in single-frame and multi-frame settings \cite{zhang2017beyond, zhang2018ffdnet, tassano2019dvdnet, tassano2020fastdvdnet, claus2019videnn}. Real low-light RAW data revealed poor transfer from sRGB or additive-Gaussian training; RAW datasets made this gap explicit and standardized evaluation \cite{plotz2017benchmarking, abdelhamed2018high}. Camera/ISP-aware data generation synthesizes RAW measurements with signal-dependent noise \cite{guo2019toward, brooks2019unprocessing, zamir2020cycleisp, abdelhamed2019noise}. For dynamic scenes, RAW video datasets support supervised learning under realistic motion and gain \cite{yue2020supervised, yue2025rvideformer}. Recent method families include single-image restorers \cite{zamir2022restormer, liang2021swinir, chen2022simple, wang2022uformer}, burst fusion \cite{mildenhall2018burst, dudhane2023burstormer}, and video models that aggregate long-range context with alignment/consistency handling \cite{chan2022basicvsr++, liang2024vrt, liang2022recurrent, yue2025rvideformer, yue2020supervised}.

Modern methods typically train on aligned noisy/clean pairs. SIDD \cite{abdelhamed2018high} is a widely used single-image benchmark with diverse sensors and a high-quality ground-truth procedure. Extending to \emph{video} is harder: reproducing identical trajectories is difficult, so object-motion datasets \cite{yue2020supervised} still exhibit noticeable misalignment; remote rigs \cite{wang2021seeing, fu2023dancing, anantrasirichai2024bvi} achieve alignment but yield low-/normal-light pairs, not suited for denoising. Exposure-dependent blur further prevents capturing matched sequences at different exposures (unlike single-image low-light denoising \cite{chen2018learning}). Consequently, existing low-light video datasets \cite{chen2019seeing, monakhova2022dancing} lack motion-consistent ground truth.

To stimulate progress under realistic constraints, we organized the AIM~2025 Low-Light RAW Video Denoising Challenge and propose a new multi-device, multi-condition dataset. Using an automated rail, we recorded 10-frame RAW bursts from 14 smartphone sensors (wide, ultrawide, telephoto, and front modules across five device) under nine conditions (exposure 1/24, 1/60, 1/120\,s; illumination 1/5/10\,lx). Six controlled-lighting scenes yield 756 sequences for training/validation; a seventh is held out for testing. High-SNR references are obtained via burst averaging (200 frames for train/val; 500 for test), extending the SIDD protocol \cite{abdelhamed2018high} to video. This provides realistic noise statistics and repeatability without synthetic degradations, establishing a fair benchmark and public leaderboard.

%

This challenge is one of the AIM 2025~\footnote{\url{https://www.cvlai.net/aim/2025/}} workshop associated challenges on: high FPS non-uniform motion deblurring~\cite{aim2025highfps}, rip current segmentation~\cite{aim2025ripseg}, inverse tone mapping~\cite{aim2025tone}, robust offline video super-resolution~\cite{aim2025videoSR}, screen-content video quality assessment~\cite{aim2025scvqa}, real-world raw denoising~\cite{aim2025rawdenoising}, perceptual image super-resolution~\cite{aim2025perceptual}, efficient real-world deblurring~\cite{aim2025efficientdeblurring}, 4K super-resolution on mobile NPUs~\cite{aim20254ksr}, efficient denoising on smartphone GPUs~\cite{aim2025efficientdenoising}, efficient learned ISP on mobile GPUs~\cite{aim2025efficientISP}, and stable diffusion for on-device inference~\cite{aim2025sd}. Descriptions of the datasets, methods, and results can be found in the corresponding challenge reports.

\section{Challenge}
\label{sec:challenge}

Existing low-light video datasets include data from only a small number of sensors and usually either provide poorly aligned ground truths or opt to rely on synthetic degradation models which limits their applicability for benchmarking real-world video denoising performance. To offer a stronger benchmark, we capture a new dataset using 14 smartphone cameras and additionally obtain high-SNR references as ground truth. This dataset forms the basis of the AIM 2025 Low-Light RAW Video Denoising Challenge.

\subsection{Challenge data}

\begin{figure}[t]
    \centering
    \includegraphics[width=\linewidth]{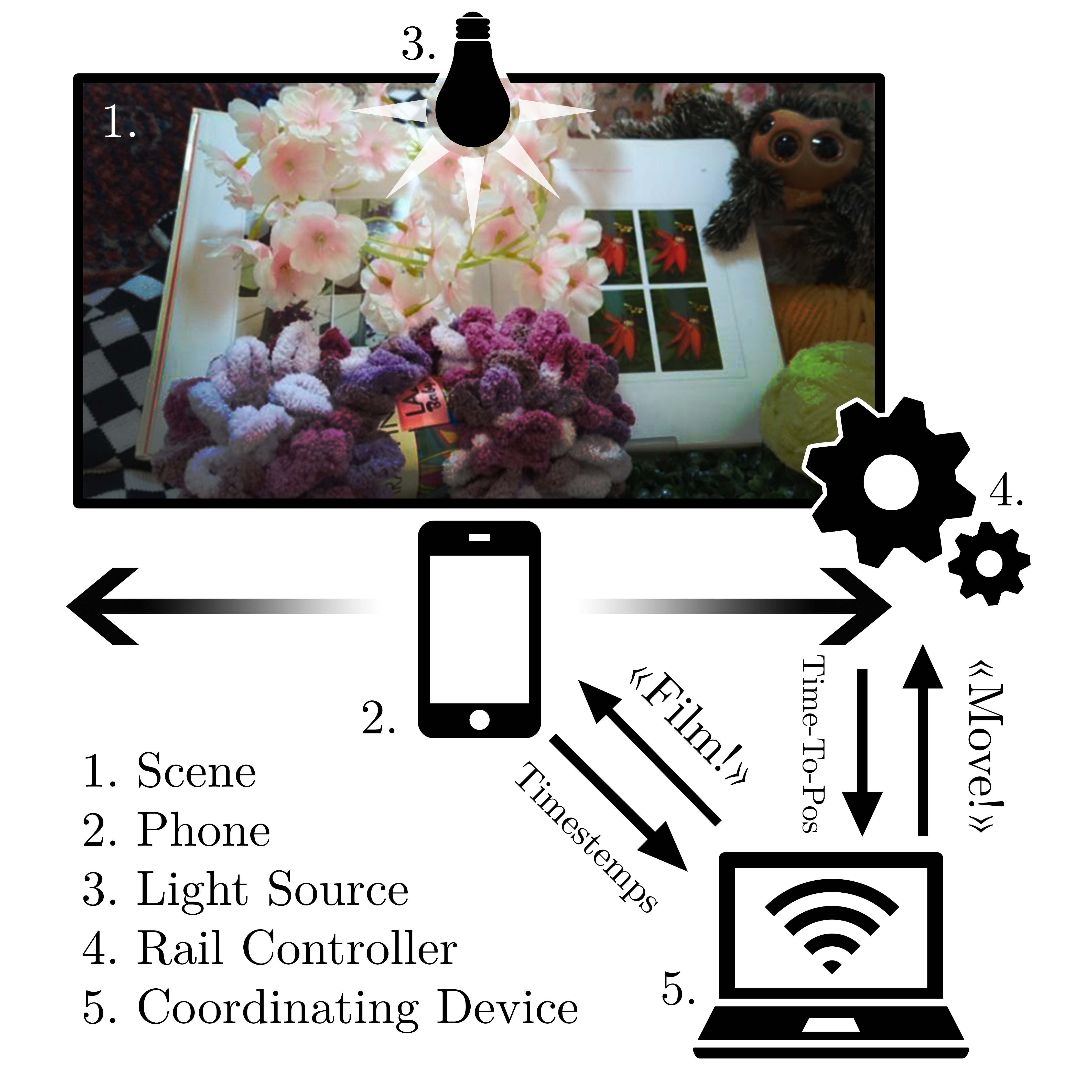}
    \caption{\textbf{Data capture setup.} The scene (1) is illuminated by a controlled light source (3) and recorded by a smartphone camera (2) mounted on a motorized rail. A rail controller (4) manages precise stage translations, while a coordinating device (5) synchronizes motion and image acquisition via wireless commands. This setup ensures repeatable, precisely aligned captures under controlled illumination, enabling consistent data collection across sensors and capture conditions.}
    \label{fig:scene-setup}
\end{figure}

For data capturing purposes we construct an automated data acquisition pipeline. The schematic of our setup is shown in Fig. \ref{fig:scene-setup}. We rigidly mount each smartphone on a motorized linear translation stage. Both stage motion and camera triggering are controlled from NVIDIA Jetson Nano. For image acquisition we modified the Pani~\cite{chugunov2024light} RAW-Burst capturing app by combining it with libsoftwareasync~\cite{AnsariSoftwareSyncICCP2019}, which enabled us to capture RAW bursts with arbitrary camera settings and full remote control. We capture the data in a stop-motion manner: move the phone to a location, capture a frame, move to the next location, capture a frame, and so on until the sequence is complete. 

We use a photo box and a non-flickering LED light for fine-grained control over scene illumination. We use a large assortment of both textured and textureless objects such as toys, books, fake flowers, etc. to ensure a wide variety of content in the captured scenes. We set the illuminance of each scene capture to 1~lx, 5~lx, or 10~lx which we verify using a luxmeter.

\begin{figure}
    \centering
    \includegraphics[width=0.95\linewidth]{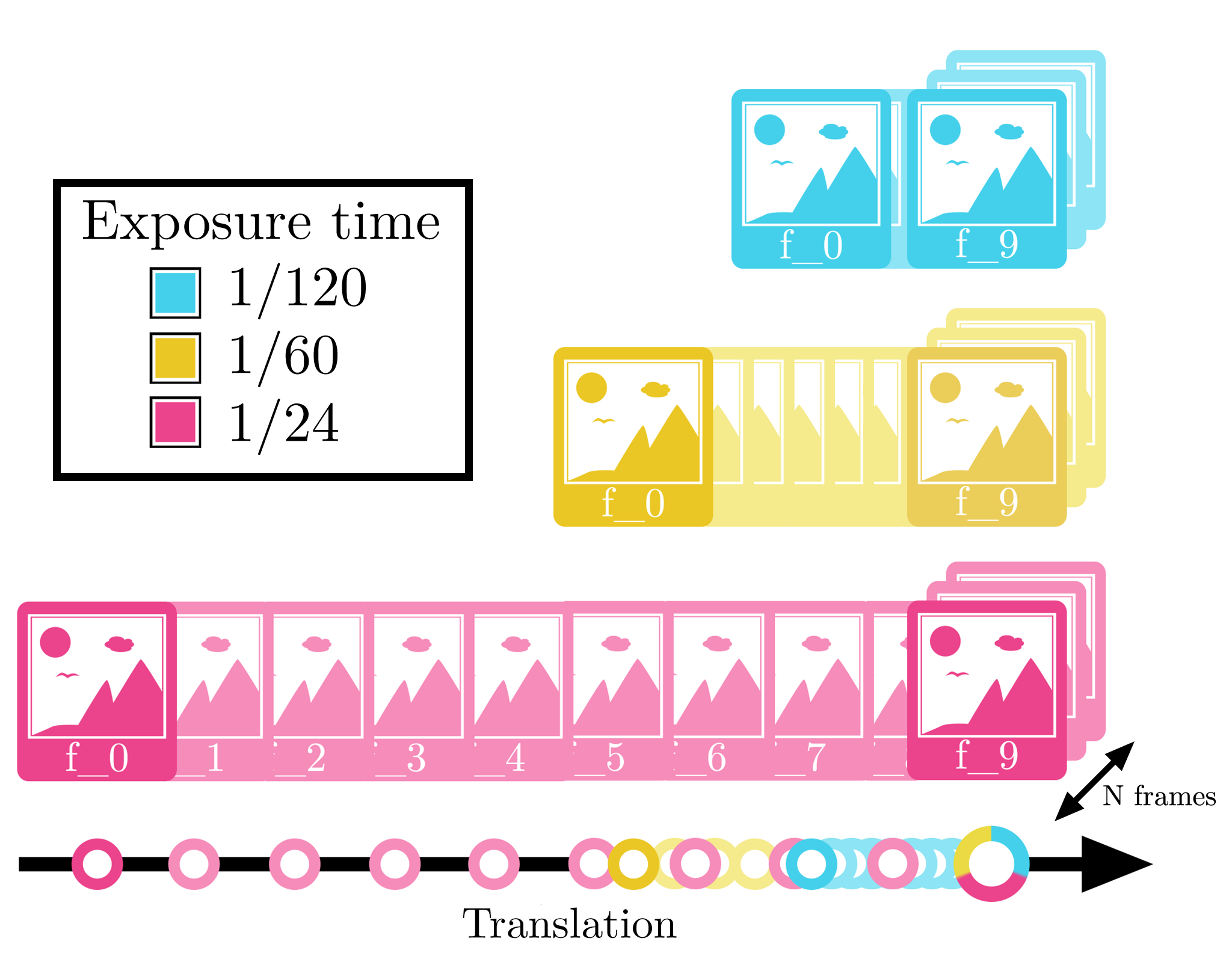}
    \caption{\textbf{Capture protocol.} Different capture settings are emulated by varying both the exposure time and the amount of motion between frames. The translation distance between consecutive frames is scaled linearly with the effective ‘frame rate’ of the capture. At the final position, an additional large burst is recorded for ground-truth calculation.}
    \label{fig:seq-capture}
\end{figure}

Three frame-rate conditions (24 fps, 60 fps, 120 fps) are emulated by changing the exposure times and by translating the stage with different step sizes between successive shots, for example, the displacement for the 1/24~s setting was five times larger than for 1/120~s, matching a realistic trade-off between the exposure time and motion magnitude (see Fig. \ref{fig:seq-capture}). Sensor ISO is adjusted for each (lux, exposure) pair so that the brightness of all the sequences is roughly the same (Tab. \ref{tab:capture-settings}).

\begin{table}[h]
  \centering
  \caption{ISO settings for each illuminance (lux) and exposure-time combination used during data capture.}
  \label{tab:capture-settings}
  \begin{tabular}{c c c c}
    \toprule
    \textbf{Illuminance} & \textbf{1/24 s} & \textbf{1/60 s} & \textbf{1/120 s} \\
    \midrule
    10 lx & 800  & 2\,000 & 4\,000 \\
     5 lx & 1\,250 & 3\,125 & 6\,250 \\
      1 lx & 2\,000 & 5\,000 & 10\,000 \\
    \bottomrule
  \end{tabular}
\end{table}

Each capture produces a 10-frame RAW sequence. To obtain a high-SNR reference for the target view (final frame of the sequence), we record a burst of additional frames and average them following the SIDD \cite{abdelhamed2018high} benchmark. For training/validation, we capture 200-shot bursts per sequence, whereas for testing we use 500-shot bursts to yield ground truths with even lower residual noise.

We collect our dataset using 14 camera modules across five smartphones: Samsung Galaxy Z Fold4, Google Pixel 5a, Google Pixel 7 Pro, Samsung Galaxy S20, and POCO X3 Pro; covering wide, ultrawide, telephoto, and front cameras (Tab. \ref{tab:devices}). This variety, together with the controlled motion, light levels, and burst-derived references, yields a rigorous benchmark for low-light RAW video denoising.

\begin{figure}[t]
  \captionsetup[subfigure]{labelformat=empty}
  \centering

  \centering
  \begin{subfigure}[b]{0.15\textwidth}
    \centering
    \includegraphics[width=\linewidth]{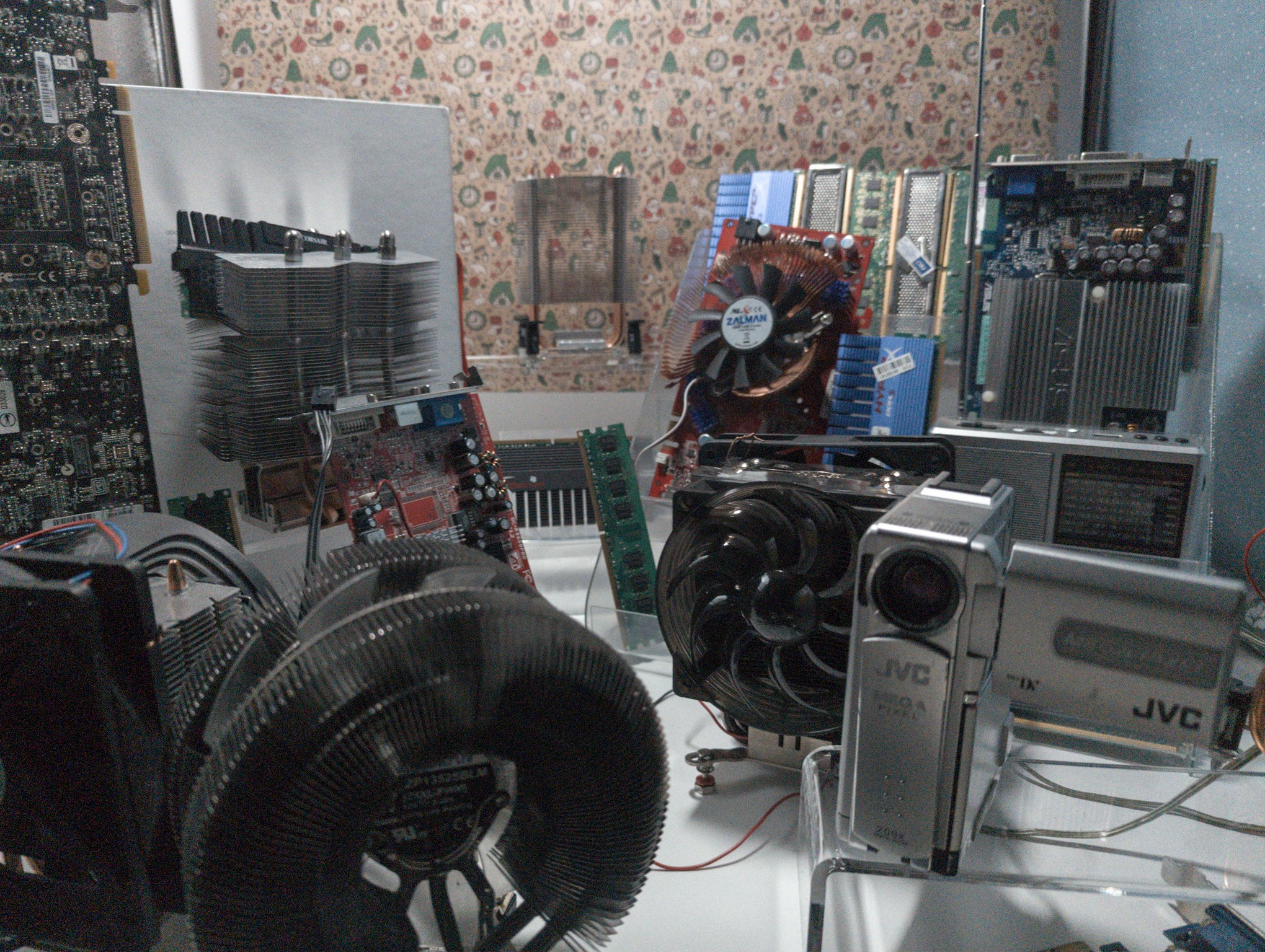}
    \caption{\texttt{hardware}}
  \end{subfigure}\hfill
  \begin{subfigure}[b]{0.15\textwidth}
    \centering
    \includegraphics[width=\linewidth]{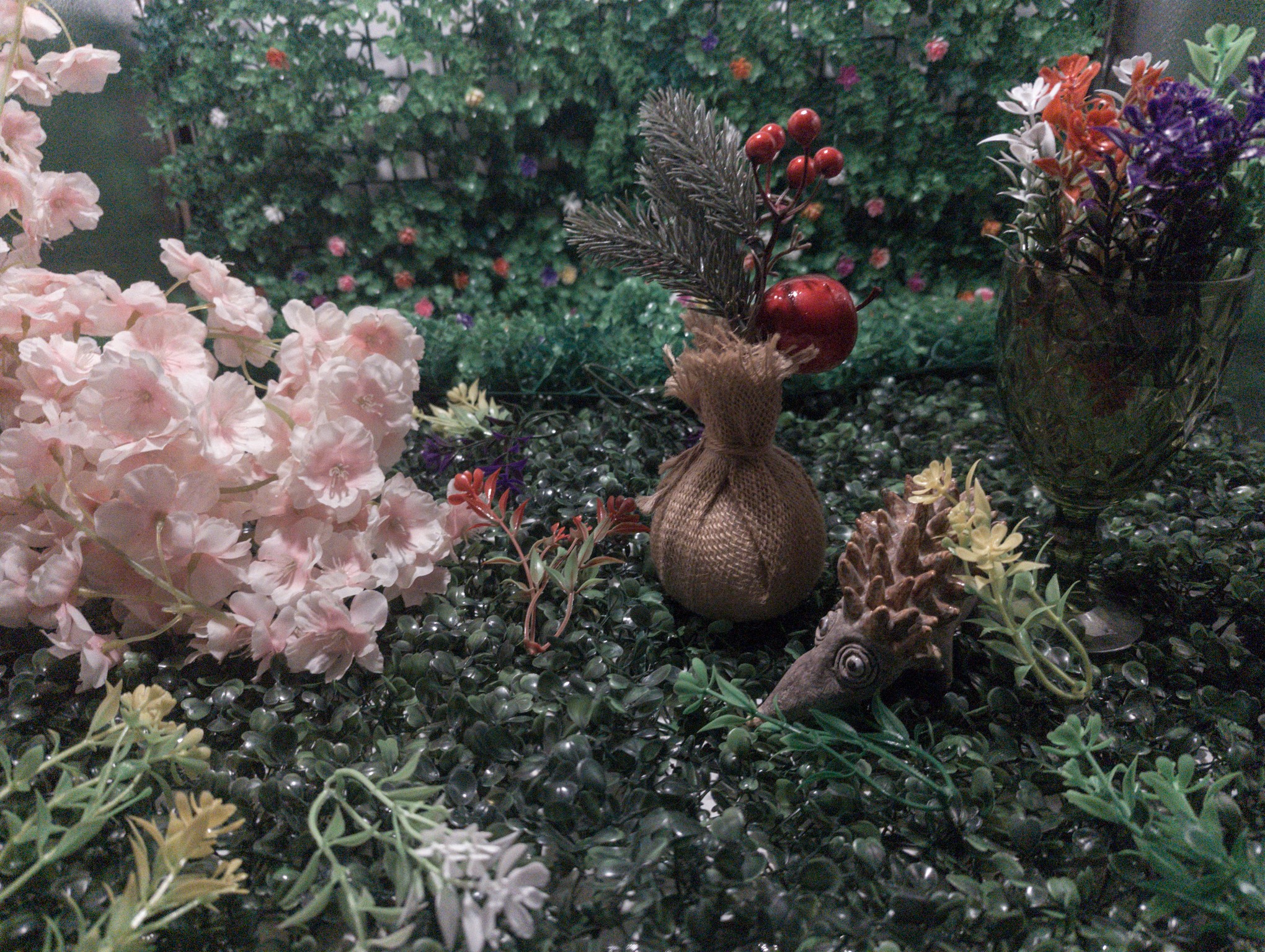}
    \caption{\texttt{spring}}
  \end{subfigure}\hfill
  \begin{subfigure}[b]{0.15\textwidth}
    \centering
    \includegraphics[width=\linewidth]{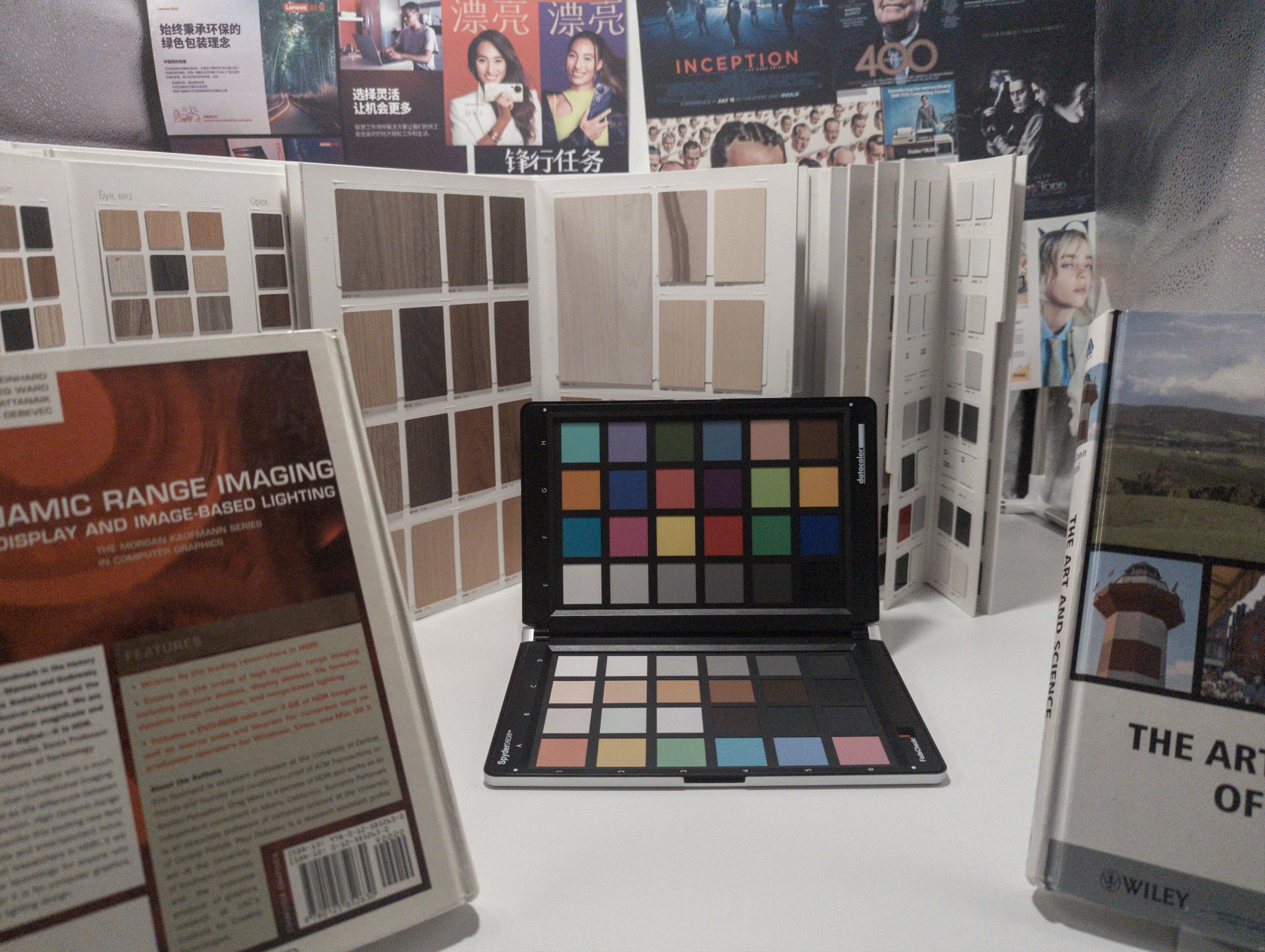}
    \caption{\texttt{color}}
  \end{subfigure}

  \vspace{4pt} 

  \begin{subfigure}[b]{0.15\textwidth}
    \centering
    \includegraphics[width=\linewidth]{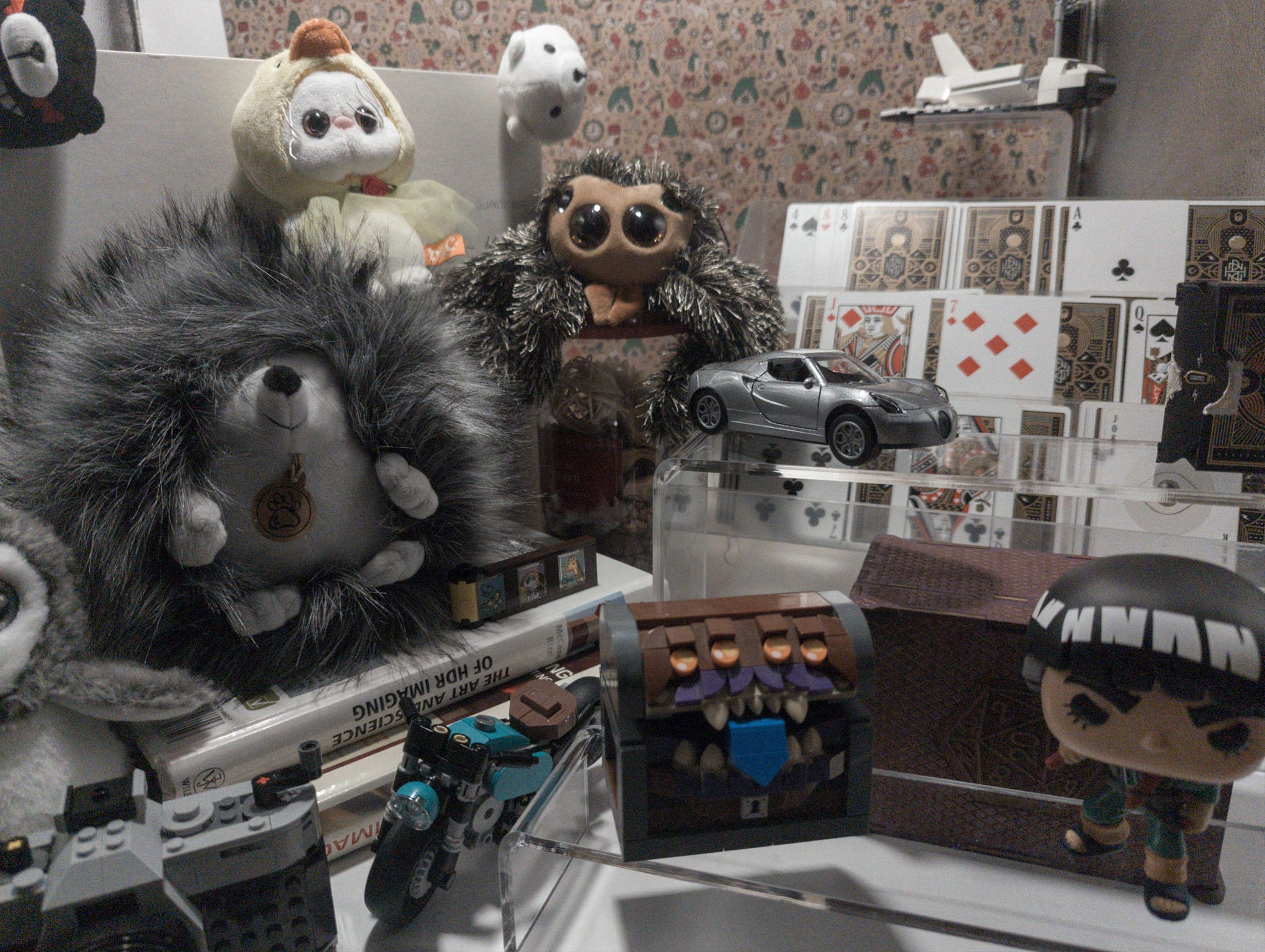}
    \caption{\texttt{toys}}
  \end{subfigure}\hfill
  \begin{subfigure}[b]{0.15\textwidth}
    \centering
    \includegraphics[width=\linewidth]{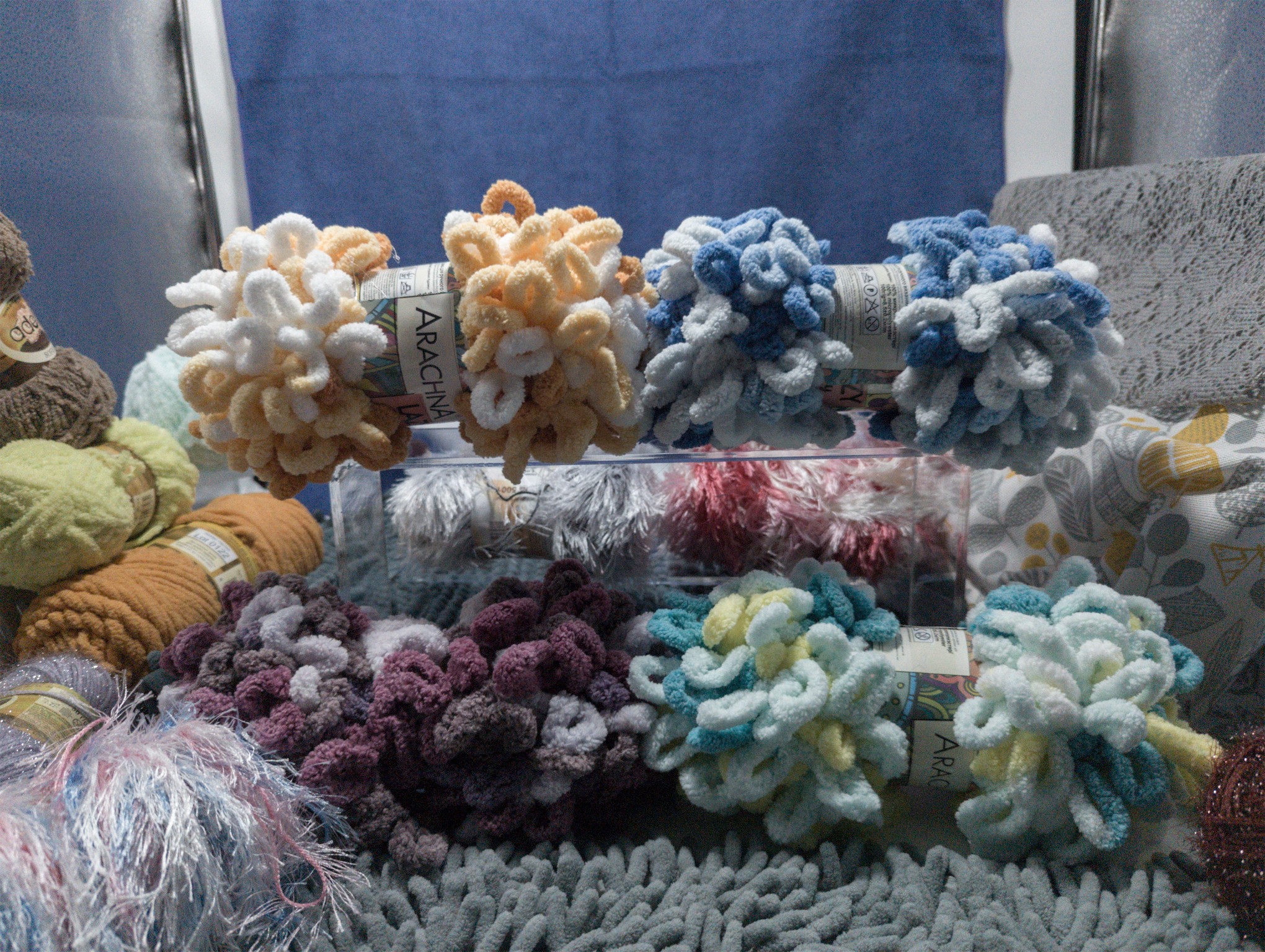}
    \caption{\texttt{yarn}}
  \end{subfigure}\hfill
  \begin{subfigure}[b]{0.15\textwidth}
    \centering
    \includegraphics[width=\linewidth]{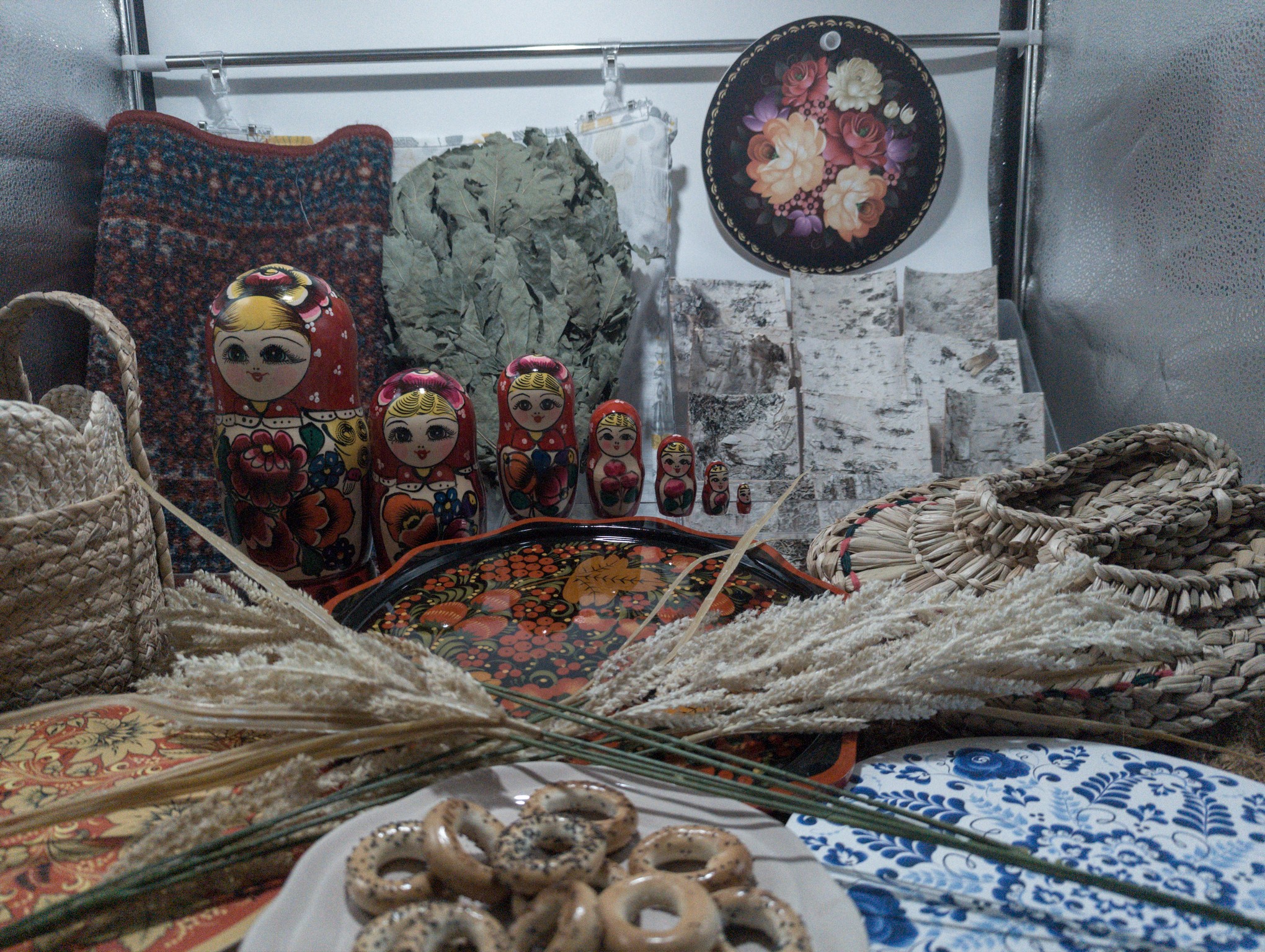}
    \caption{\texttt{folk}}
  \end{subfigure}
  
  \caption{\textbf{Dataset scene overview.} We collected 6 distinct scenes with varying content. We included objects with varying levels of details. Pictures shown were taken using the main camera of Google Pixel 7 Pro at 1/24\,s exposure and with 10\,lx of illumination. Viewer is advised to zoom in.}
  \label{fig:dataset_scenes}
\end{figure}

In total we create 6 capture scenes showcased in Fig. \ref{fig:dataset_scenes}, and capture a total of 756 RAW video sequences (=~14 sensors $\times$ 9 conditions $\times$ 6 scenes) to be used for method training.
An additional seventh scene is captured for testing, the contents of this scene are constructed out of the same objects but with slight changes to prevent overfitting; exactly one (lux, exposure) condition per sensor was selected at random and subsequently captured using our automated setup.


\begin{table}[t]
  \centering
  \caption{\textbf{Camera sensor list.} The dataset was captured using 14 camera sensors from 5 smartphones. For each sensor, we also report the maximum supported ISO, as some sensors could not reach the designated ISO values listed in Tab.~\ref{tab:capture-settings}.}
  \label{tab:devices}

\begin{tabular}{cccc}
  \toprule
  \textbf{Device} & \textbf{Camera} & \textbf{Max ISO} & \textbf{Resolution} \\ \midrule
  \multirow{3}{*}{\dev{Samsung}{Galaxy\,Z Fold4}} & Wide       & 1600  & 4080$\times$3060 \\
                                                  & Ultrawide  & 10000 & 4080$\times$3060 \\
                                                  & Telephoto  & 10000 & 3648$\times$2736 \\
  \cdashlinelr{1-4}
  \multirow{3}{*}{\dev{Google}{Pixel 5a}}         & Wide       & 7109  & 4032$\times$3024 \\
                                                  & Ultrawide  & 9208  & 4032$\times$3022 \\
                                                  & Front      & 10000 & 3280$\times$2464 \\
  \cdashlinelr{1-4}
  \multirow{4}{*}{\dev{Google}{Pixel 7 Pro}}      & Wide       & 10000 & 4080$\times$3072 \\
                                                  & Ultrawide  & 10000 & 4080$\times$3072 \\
                                                  & Telephoto  & 1143  & 4080$\times$3072 \\
                                                  & Front      & 3918  & 3440$\times$2448 \\
  \cdashlinelr{1-4}
  \multirow{2}{*}{\dev{Samsung}{Galaxy S20}}      & Wide       & 9993  & 4000$\times$3000 \\
                                                  & Ultrawide  & 10000 & 4000$\times$3000 \\
  \cdashlinelr{1-4}
  \multirow{2}{*}{\dev{POCO}{X3 Pro}}             & Wide       & 10000 & 4000$\times$3000 \\
                                                  & Front      & 10000 & 2592$\times$1940 \\ 
  \bottomrule
\end{tabular}

\end{table}

\subsection{Challenge organization}

The challenge was set up using CodaBench~\cite{codabench} platform. 
It consisted of two development stages and a final testing stage. For the first development stage, \texttt{hardware} and \texttt{spring} scenes were released for training, while a portion of \texttt{yarn} scene was used as validation. For the second stage we released the rest of \texttt{yarn} sequences and full \texttt{color} and \texttt{toys} scenes, leaving \texttt{folk} to be used as validation. For each validation subset we use the same 14 condition-sensor pairs as in the final test data.
The data was released in the form of \texttt{.npz} compressed dictionaries each containing the 10-frame noisy video sequence along with 20 additional independent noise realizations of the last frame and higher-SNR GT obtained via a pipeline similar to SIDD \cite{abdelhamed2018high}. All images were transformed to [0, 1] value range by removing black level and mapping white level to 1, the black/white level values were retrieved from capture metadata.

\subsection{Evaluation}

For evaluation, we compute PSNR and SSIM \cite{wang2003multiscale} on the linear, mosaicked RAW data. Each submission is ranked according to both metrics, and the average of the two ranks determines the final leaderboard position.

\subsection{Submissions}

For each validation / test sequence the participants were tasked with submitting the center 1024$\times$1024 crop of the denoised 10th frame, preserving both the linear RAW format and the Bayer pattern.
In order to be eligible for inclusion in the final challenge report the teams were required to send the code that reproduces the final submissions.


\section{Challenge results}
In this section we showcase main challenge results on test data and provide additional method performance analysis using held out portion of \texttt{folk} scene.

\subsection{Test data results}

\begin{table}
  \centering
  \caption{\textbf{Results of the AIM 2025 Low-light RAW Video Denoising Challenge.} PSNR and SSIM are computed on the 14 sequences of the private test set. Final rankings are determined by averaging the ranks from both metrics.}
  \label{tab:results}
  \begin{tabular}{ l c c c c c c }
\toprule
\multirow{2}[2]{*}{Method} & \multirow{2}[2]{*}{\dev{Multi-}{frame}} & \multicolumn{2}{c}{Metric} & \multirow{2}[2]{*}{\dev{Final}{rank}} \\ \cmidrule(lr){3-4} 
\multicolumn{1}{c}{}       &                & PSNR $\uparrow$  & SSIM $\uparrow$ &        \\ \midrule
\rowcolor{gold}
SNU-ISPL                   & $\checkmark$   & \textbf{48.32} & \textbf{0.9879} & 1      \\
\rowcolor{silver}
XJAI                       & $\checkmark$   & \underline{48.19} & \underline{0.9865} & 2      \\
\rowcolor{bronze}
AxeraAI                    & $\times$       & 46.52 & 0.9814 & 3      \\
VMCL-ISP                   & $\times$       & 45.76 & 0.9682 & 4      \\ \midrule
UNet w/ Attn               & $\checkmark$   & 45.72 & 0.9797 & ---    \\
UNet                       & $\times$       & 43.52 & 0.9691 & ---    \\
Noisy                      & ---            & 36.06 & 0.8093 & ---    \\ \bottomrule
\end{tabular}
\end{table}

From 52 registered participants, 4 teams successfully entered the final phase, submitting valid results, executable code, and method descriptions. Tab. \ref{tab:results} summarizes the final results, reporting PSNR and SSIM on the private test set, while Fig. \ref{fig:qualitative} presents qualitative examples on two representative sequences.

The best results were achieved by SNU-ISPL, closely followed by XJAI, with a significant margin separating these two multi-frame solutions from the remaining single-frame approaches. This performance gap highlights the substantial benefits of leveraging temporal information for low-light RAW video denoising. Both top-ranked teams exploited multiple noisy frames, whereas the other two submissions relied on single-frame processing.

SNU-ISPL’s winning method, DarkVRAI, combines a selective burst-scan mechanism with capture-condition conditioning for both alignment and denoising, achieving superior temporal aggregation. XJAI adopts an efficient hierarchical transformer architecture with progressive refinement, processing sequences of nine frames with noise variation to improve robustness. The third-ranked team, AxeraAI, focuses on convolutional models, employing a high-performing single-frame NAFNet \cite{chen2022simple} baseline. VMCL-ISP builds on their prior YOND \cite{yond} framework, emphasizing data-centric noise modeling with a variance-stabilizing transform and Restormer-based denoiser \cite{zamir2022restormer}.

For reference, we also include two baselines: a single-image UNet \cite{ronneberger2015u} trained with a Noise2Noise \cite{lehtinen2018noise2noise} objective, and its channel-attention variant for noisy-frame fusion. Consistent with the challenge results, these baselines further confirm the advantage of multi-frame processing for RAW video denoising.

\subsection{Extended results on validation data}

We further study the \textit{cross-sensor} and \textit{cross-condition} performance of the challenge methods using the complete \texttt{folk} scene, which contains data from all 14 sensors across 9 capture conditions. We select this scene in particular, because most of these sequences were unseen during the training of participants' solutions.

\subsubsection{Cross-sensor performance}

Tab.~\ref{tab:folk-per-device} reports PSNR values averaged over all capture conditions for each sensor. Among the multi-frame methods, SNU-ISPL is the most consistent, outperforming or matching competitors on 9 out of 14 sensors. Its mean PSNR is slightly lower than XJAI due to larger performance gaps on the Galaxy Z Fold4 Telephoto and POCO X3 Pro Wide cameras. We attribute part of this discrepancy to differences in ground truth quality between the validation and final test sets (200 vs.\ 500-frame averaging).
For single-frame methods, AxeraAI achieves the highest per-sensor wins, leading on 10 out of 14 sensors. The multi-frame baseline UNet with attention consistently outperforms its single-frame variant, confirming that temporal information is beneficial even for a relatively simple architecture.

\begin{table*}[t]
    \caption{\textbf{Cross-sensor performance.} We test the performance of submitted solutions on the \texttt{folk} scene and list PSNR values corresponding to each camera sensor. We list the phone model and the camera module: W, UW, T, F --- stand for Wide, Ultrawide, Telephoto, and Front camera sensors, respectively.}
    \label{tab:folk-per-device}
    \centering
    \setlength\tabcolsep{4.25pt}
\begin{tabular}{l c c c c c c c c c c c c c c}
\toprule
\multirow{2}[2]{*}{Method} & \multicolumn{3}{c}{\dev{Samsung}{Galaxy\,Z Fold4}} & \multicolumn{3}{c}{Google Pixel 5a} & \multicolumn{4}{c}{Google Pixel 7 Pro} & \multicolumn{2}{c}{\dev{Samsung}{Galaxy S20}} & \multicolumn{2}{c}{\dev{POCO}{X3 Pro}} \\
\cmidrule(lr){2-4} \cmidrule(lr){5-7} \cmidrule(lr){8-11} \cmidrule(lr){12-13} \cmidrule(lr){14-15}
 & W & UW & T & W & UW & F & W & UW & T & F & W & UW & W & F \\
\midrule
SNU-ISPL & \textbf{52.36} &	\textbf{49.44} &	46.15 &	            \textbf{43.16} &	\textbf{46.69} &	\textbf{45.56} &	 45.40 &	        \textbf{45.18} &	53.00 &	\textbf{45.79} &	52.60 &	    \textbf{45.19} &	45.53 &	            \textbf{48.39} \\
XJAI     & 52.04 &	        \textbf{49.44} &	\textbf{47.60} &	43.01 &	             46.66 &	         45.02 &	         \textbf{45.45} &	45.12 &	\textbf{53.30} &	    45.47 &	            \textbf{52.73} &	45.07 &	          \textbf{47.01} &	48.36 \\
\cdashlinelr{1-15}
VMCL-ISP & 46.92 &	        \textbf{48.07} &	45.17 &	            41.01 &	            44.51 &	            43.44 &	            44.43 &	            \textbf{43.19} &	52.48 &	                         43.45 &	            \textbf{52.51} &	\textbf{43.46} &	46.19 &	            46.35 \\
AxeraAI  & \textbf{48.63} &	47.92 &	            \textbf{46.04} &	\textbf{41.19} &	\textbf{44.92} &	\textbf{43.49} &	\textbf{44.69} &	42.96 &	            \textbf{52.51} &	            \textbf{43.54} &	51.77 &	            43.32 &	            \textbf{46.24} &	\textbf{46.37} \\
\cdashlinelr{1-15}
UNet w/ Attn &     \textbf{49.45} &	\textbf{46.84} &	\textbf{45.93} &	\textbf{39.74} &	\textbf{44.03} &	\textbf{41.77} &	\textbf{42.14} &	\textbf{42.35} &	\textbf{50.66} &	\textbf{43.47} &	\textbf{49.74} &	\textbf{42.90} &	\textbf{44.32} &	\textbf{45.70} \\
UNet &      46.86 &	43.68 &	43.81 &	38.78 &	42.22 &	40.52 &	40.91 &	40.88 &	49.06 &	41.68 &	46.51 &	40.63 &	43.50 &	43.30 \\
\bottomrule
\end{tabular}
\end{table*}

\subsubsection{Cross-condition performance}

\begin{figure}
    \centering
    \includegraphics[width=0.8\linewidth]{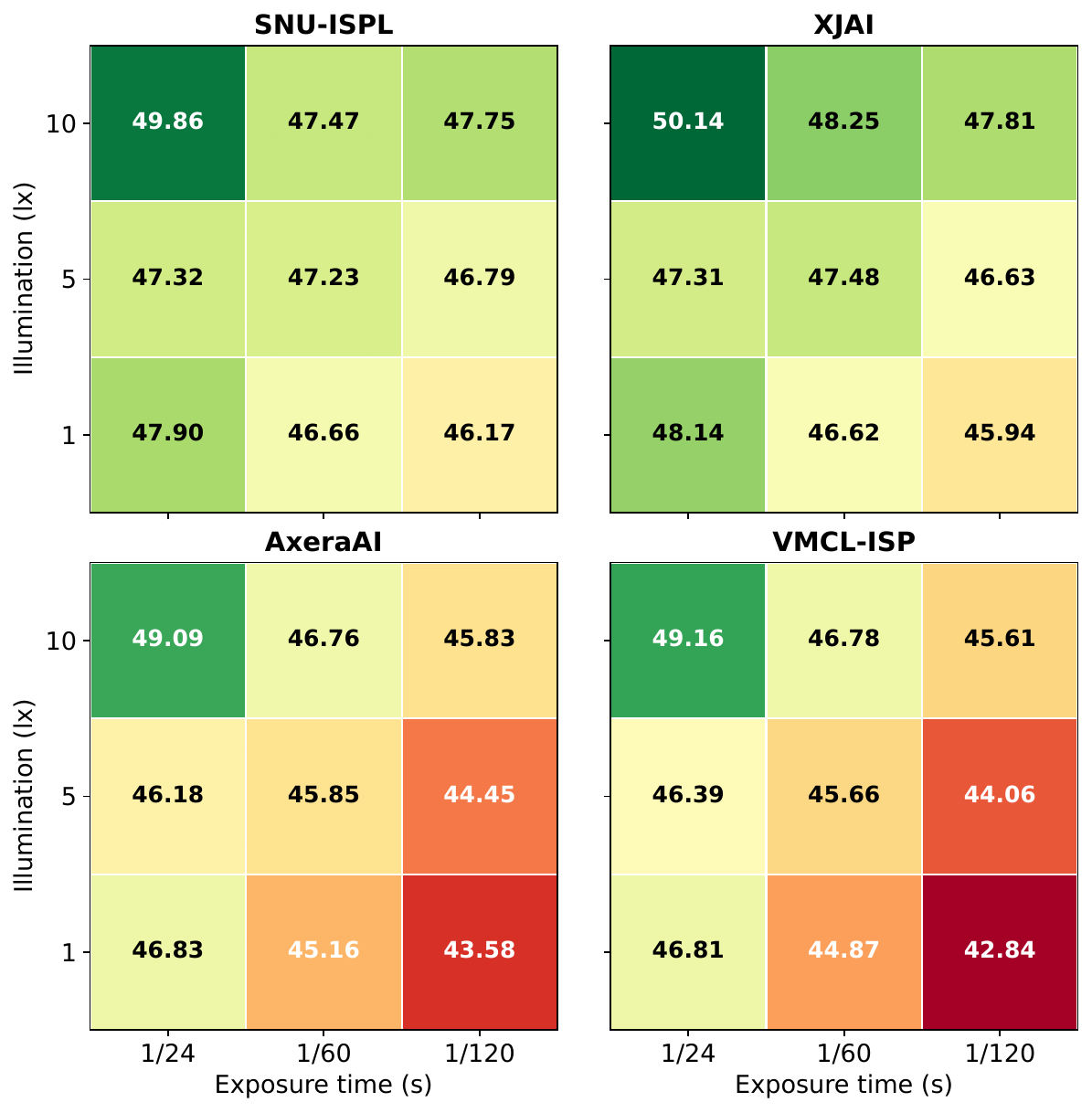}
    \caption{\textbf{Cross-condition performance.} We test the submitted solutions on the \texttt{folk} scene and plot each methods performance (PSNR values) depending on the shooting conditions, averaged by all devices.}
    \label{fig:heatmap}
\end{figure}

Fig.~\ref{fig:heatmap} shows the performance of all methods across illumination-exposure combinations on the \texttt{folk} scene. While multi-frame methods generally achieve higher PSNR, the magnitude of improvement depends on the noise level. In bright conditions (10~lx, 1/24~s), all methods perform similarly, indicating that temporal information offers limited gains when SNR is high. At shorter exposures or lower illumination, performance differences widen, and in the noisiest regime (1~lx, 1/120~s) the best multi-frame methods exceed single-frame approaches by over 3~dB.

We observe that (i) XJAI maintains more even performance across illumination levels, particularly at intermediate exposures, (ii) SNU-ISPL remains competitive across most conditions but shows slightly larger drops in some extreme settings, and (iii) single-frame methods (AxeraAI, VMCL-ISP) degrade more sharply in low-light, short-exposure regimes, reflecting the limits of spatial-only denoising in RAW video.

Overall, while mean PSNR determines leaderboard position, per-sensor and per-condition analyses highlight differences in robustness: some methods achieve higher peak performance on select sensors or conditions, whereas others offer more uniform results across the entire range of capture setups.

\newcommand{\colgap}{6pt}   
\newcommand{\vertgap}{3pt}  
\newcommand{\cropgap}{3.5pt}  

\newcommand{\tile}[3]{%
  \begin{minipage}[t]{0.15\linewidth}
    \includegraphics[width=\linewidth]{#1}\vspace{\vertgap}\\
    \includegraphics[width=.475\linewidth]{#2}%
    \hspace{\cropgap}%
    \includegraphics[width=.475\linewidth]{#3}%
  \end{minipage}%
}

\begin{figure*}[t]
  \centering
  \begin{tabular}{@{}*{5}{c@{\hspace{\colgap}}}c@{}}
    \tile{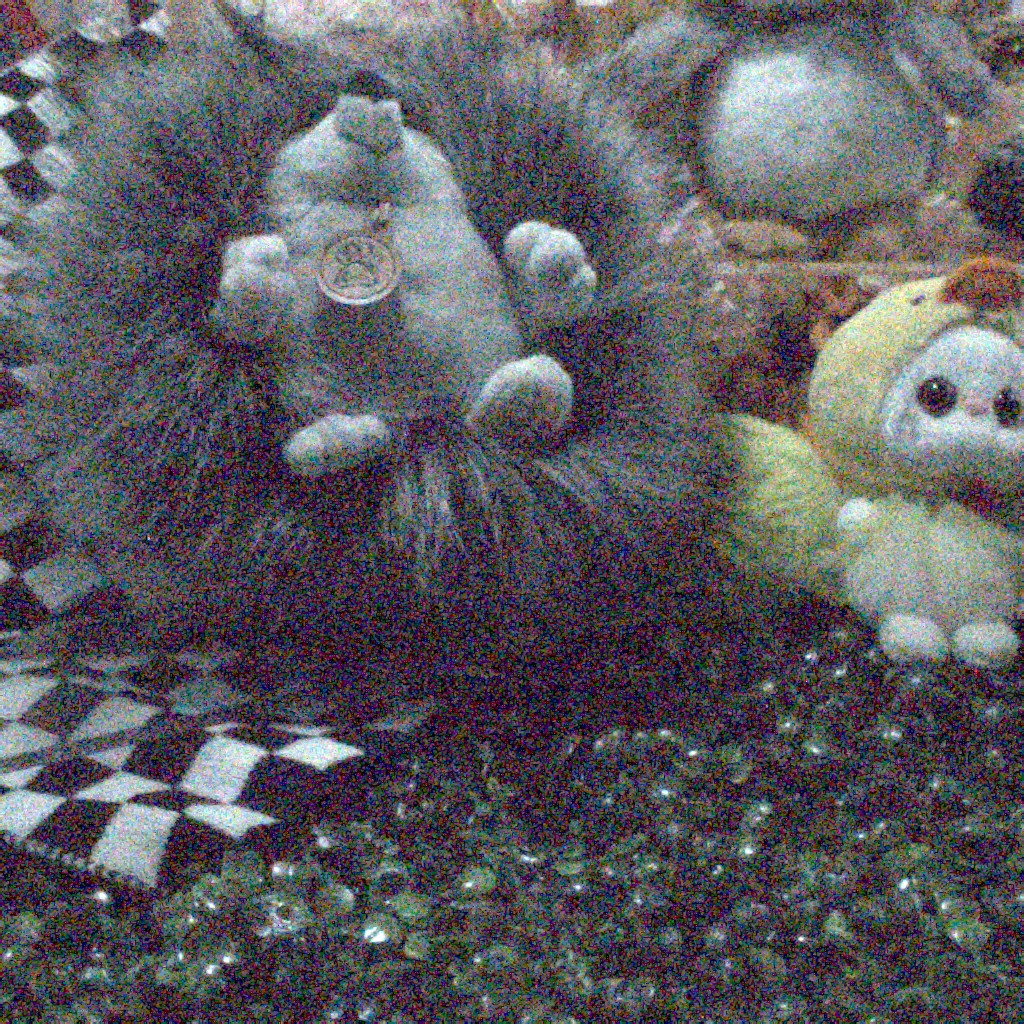}{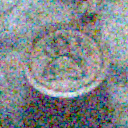}{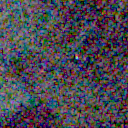} &
    \tile{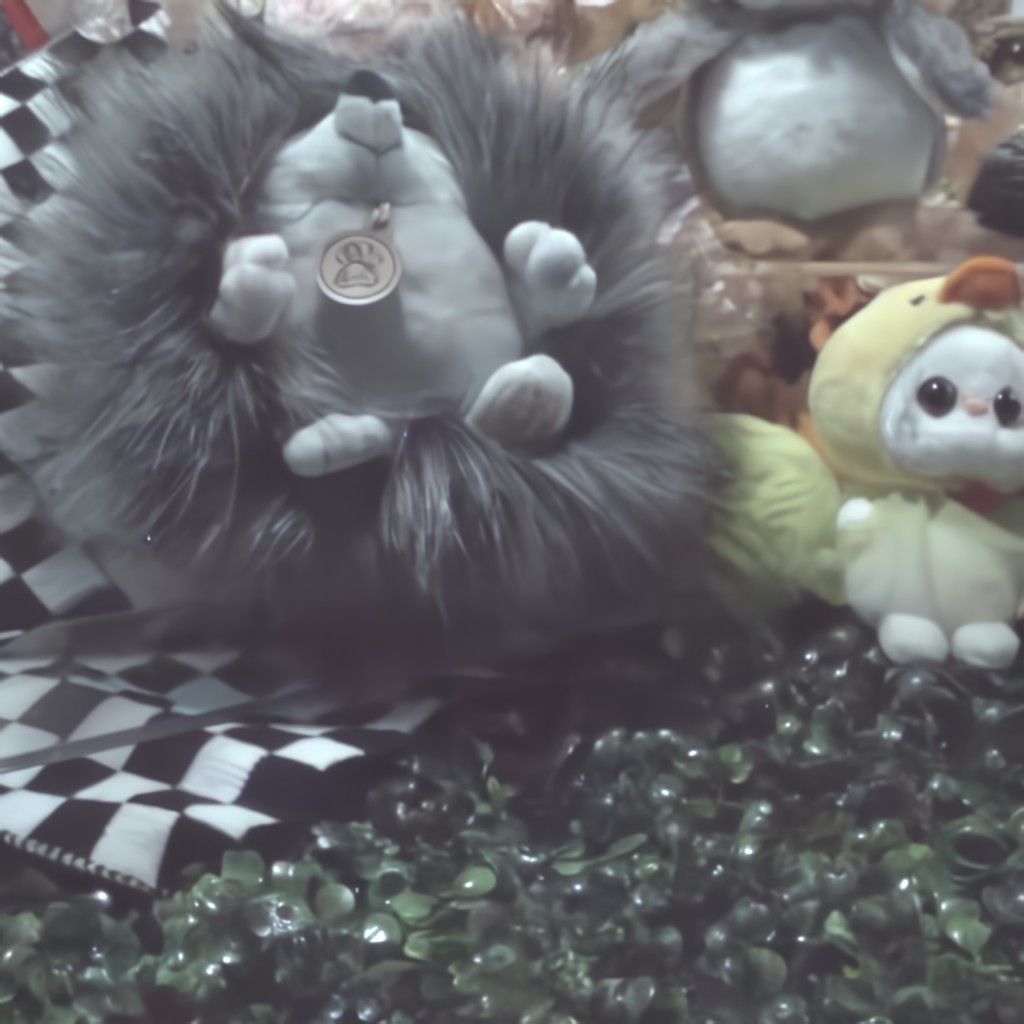}{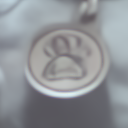}{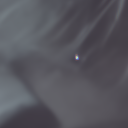} &
    \tile{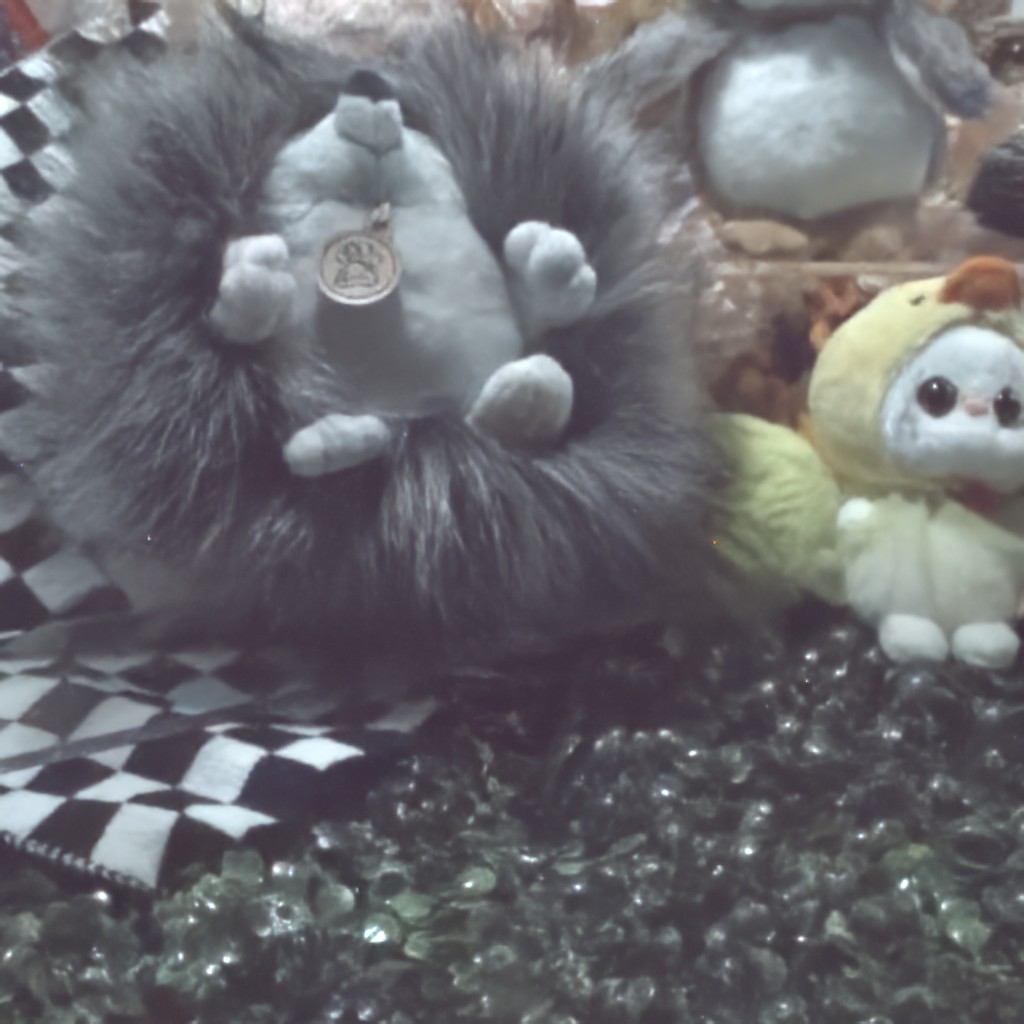}{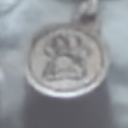}{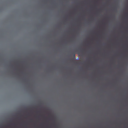} &
    \tile{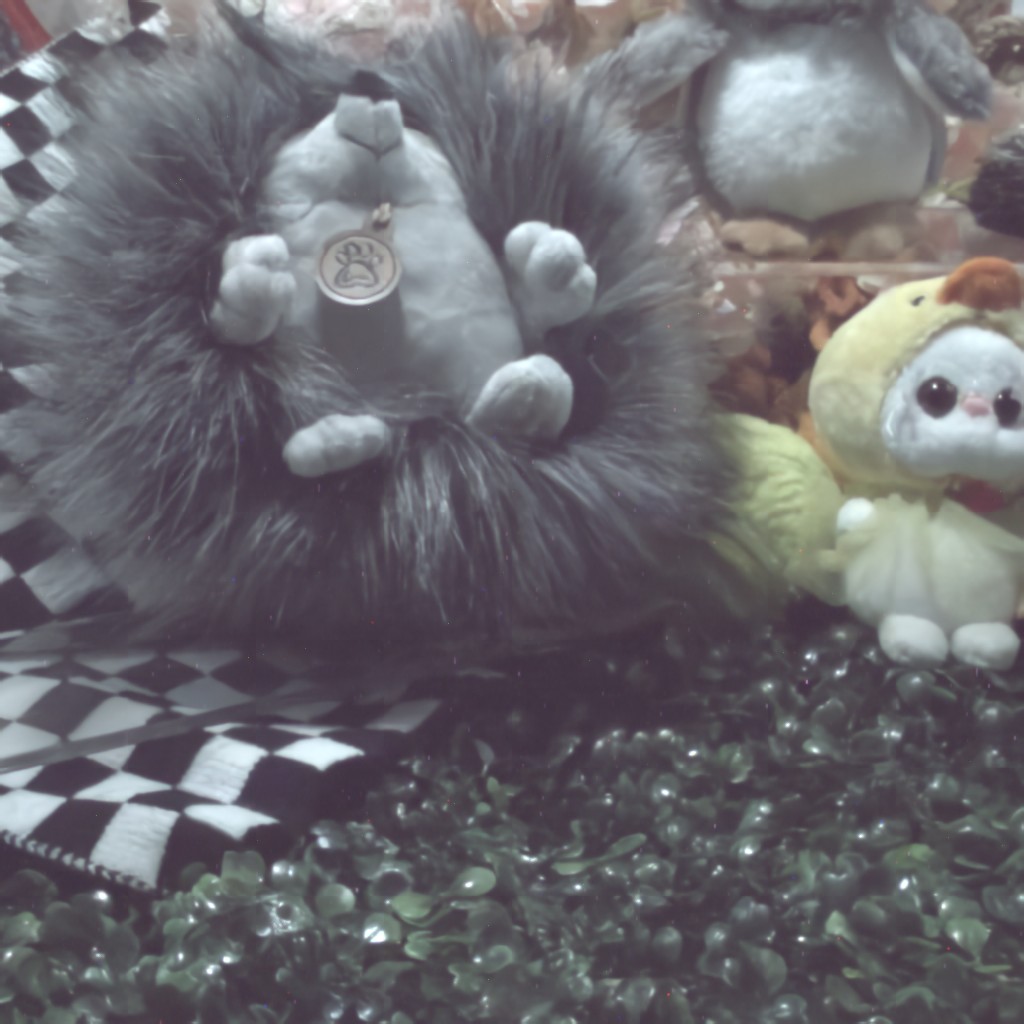}{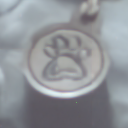}{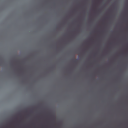} &
    \tile{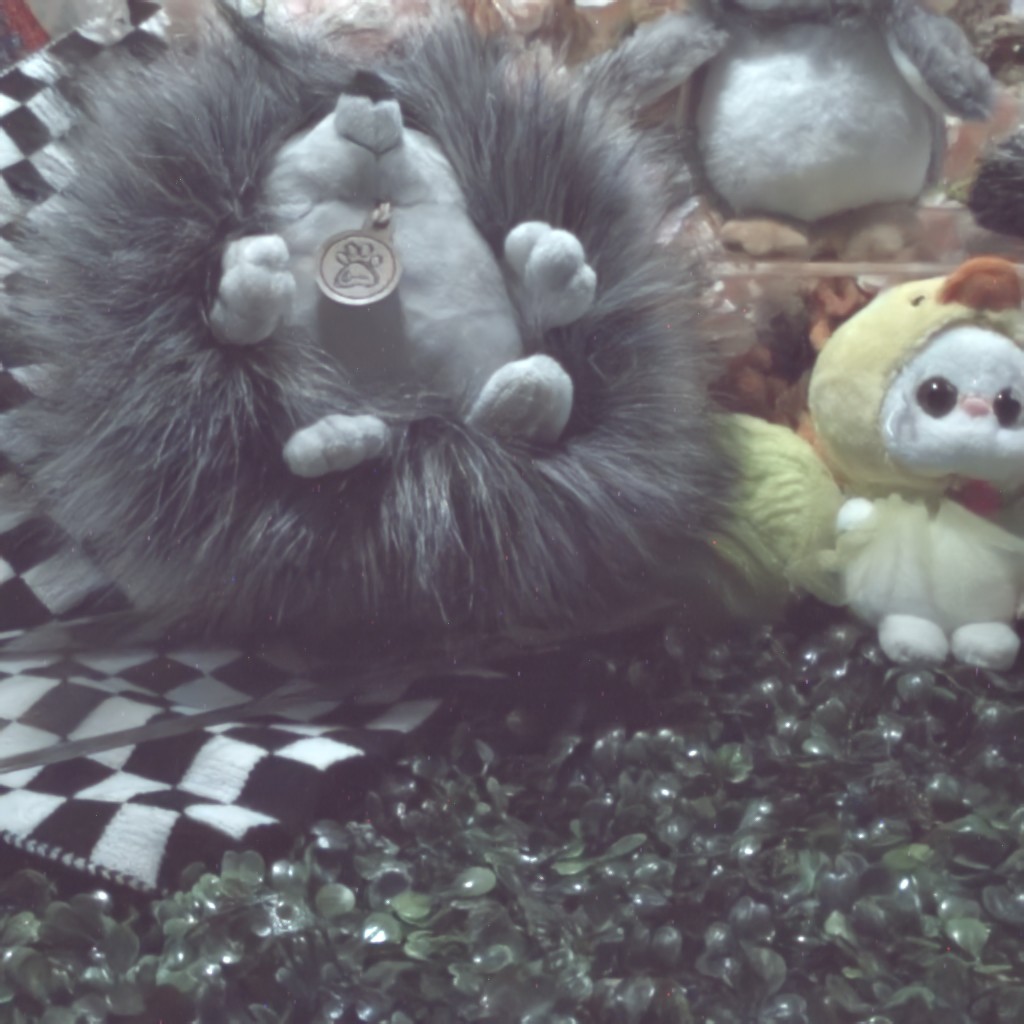}{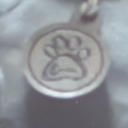}{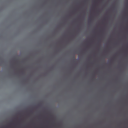} &
    \tile{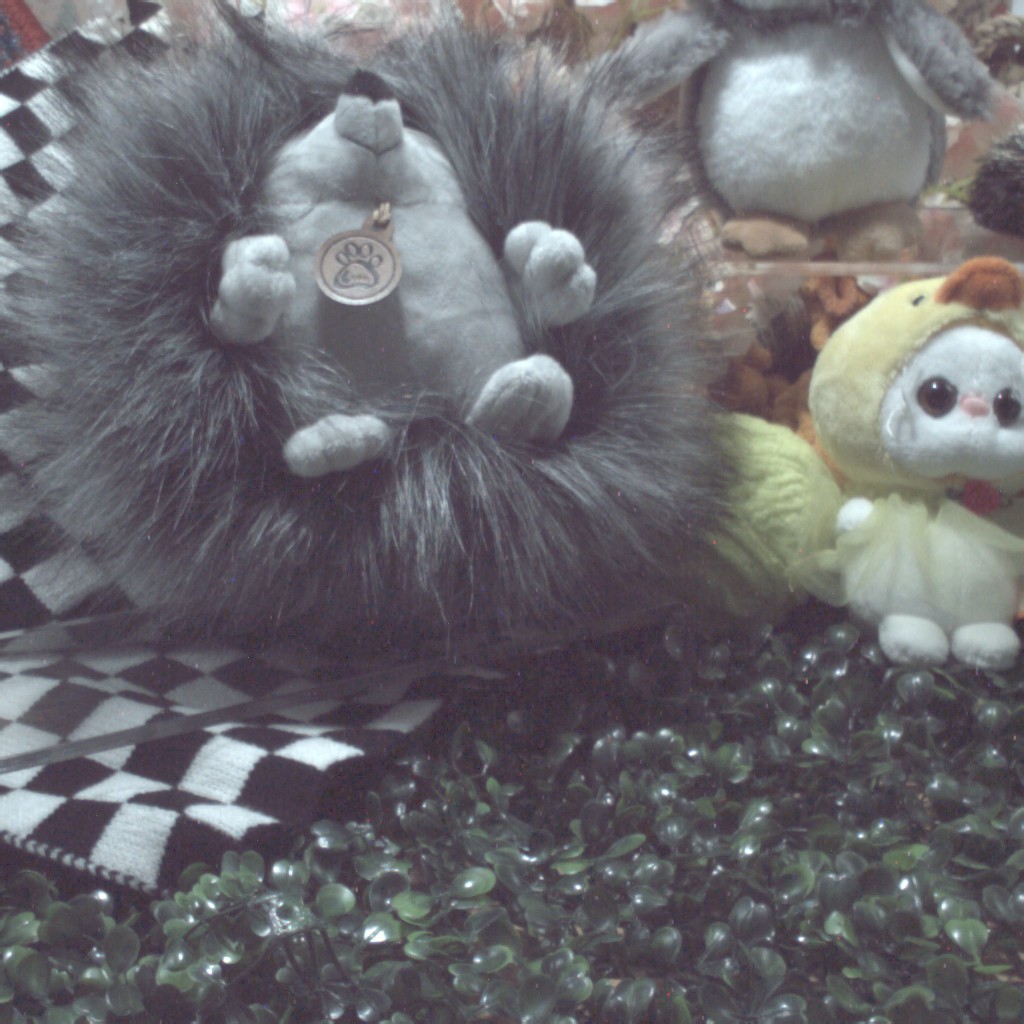}{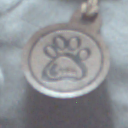}{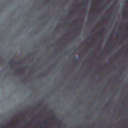} \\
    \vspace{-6pt}\\
    \tile{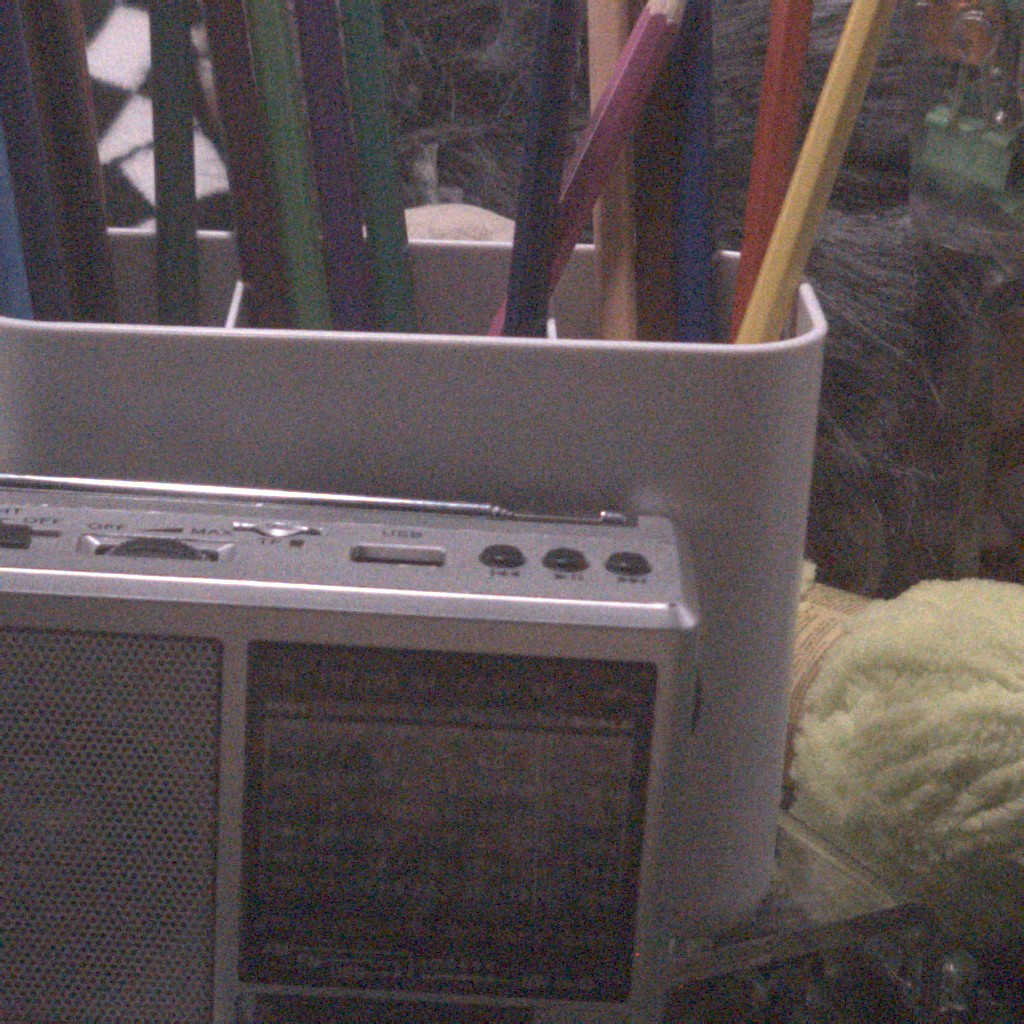}{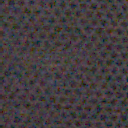}{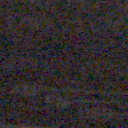} &
    \tile{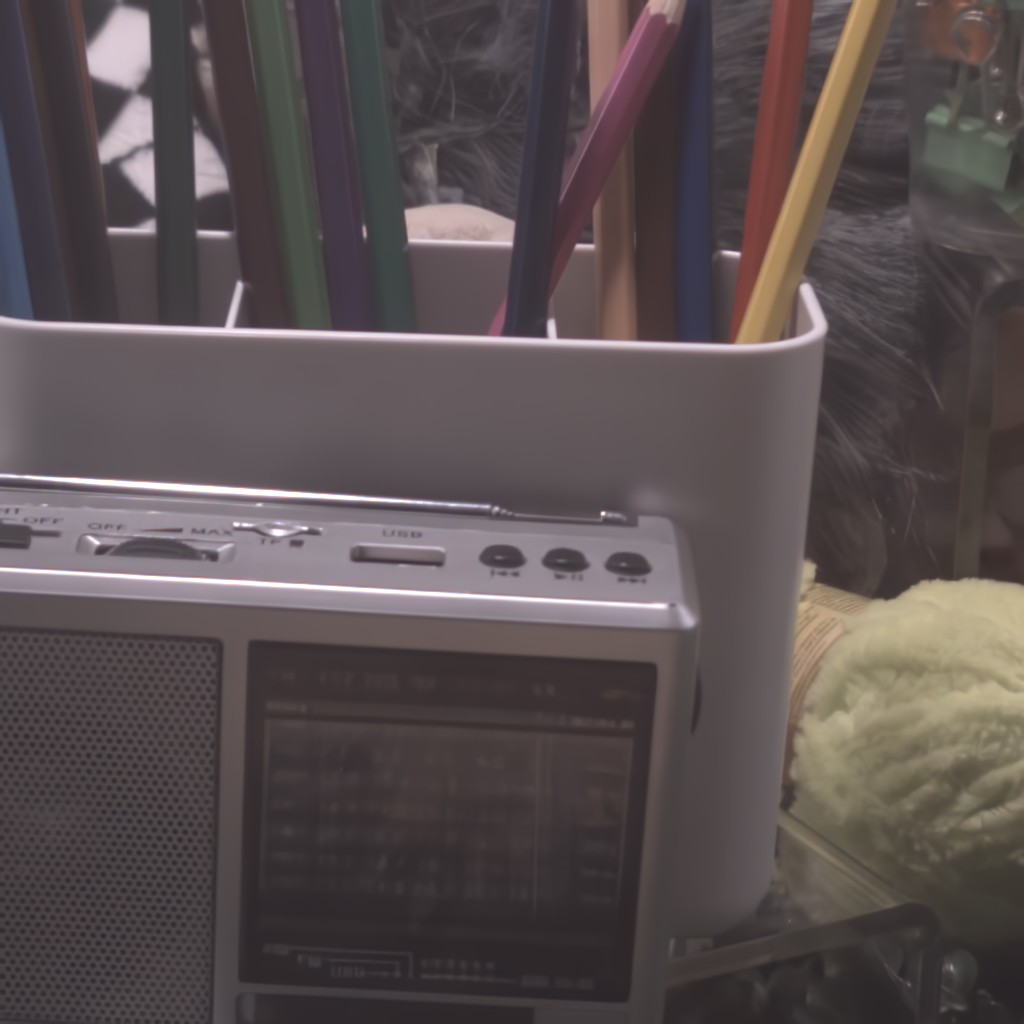}{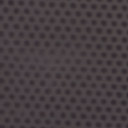}{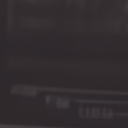} &
    \tile{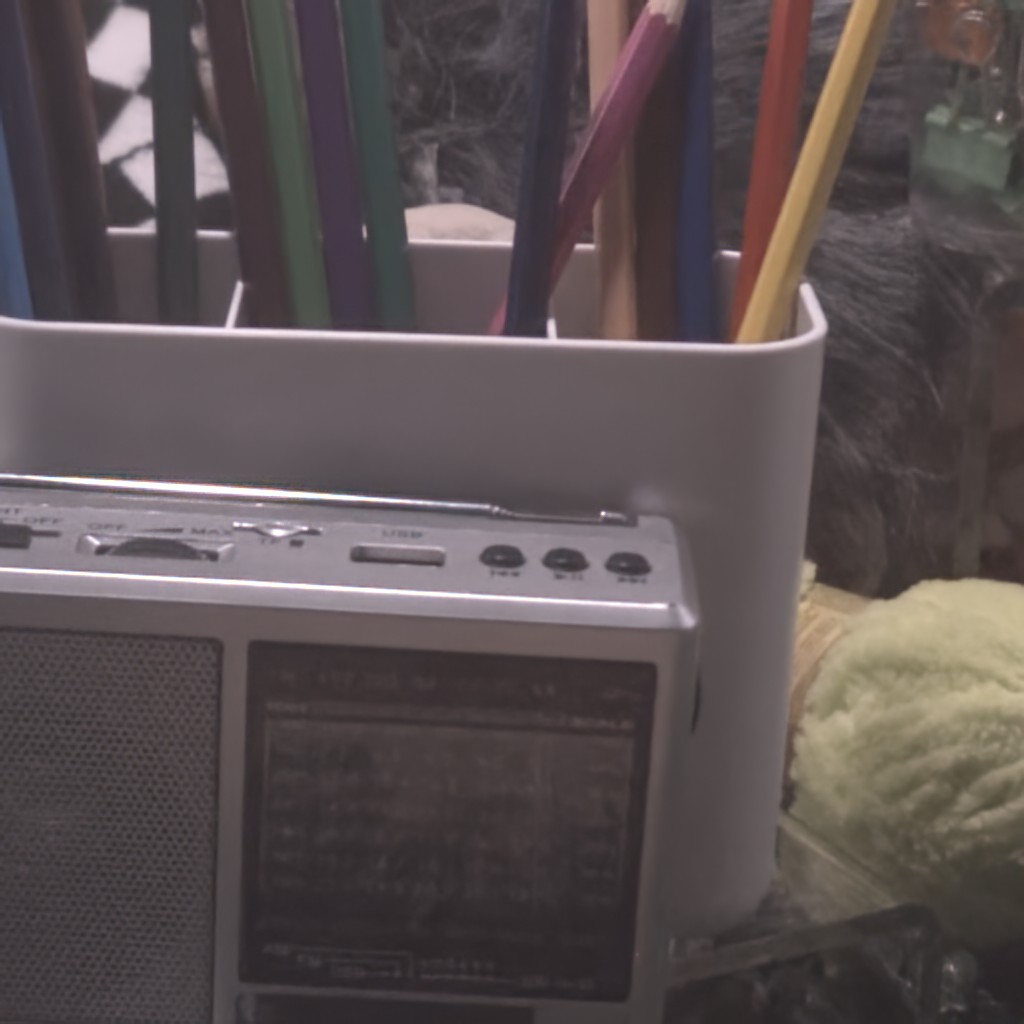}{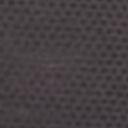}{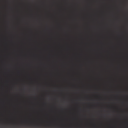} &
    \tile{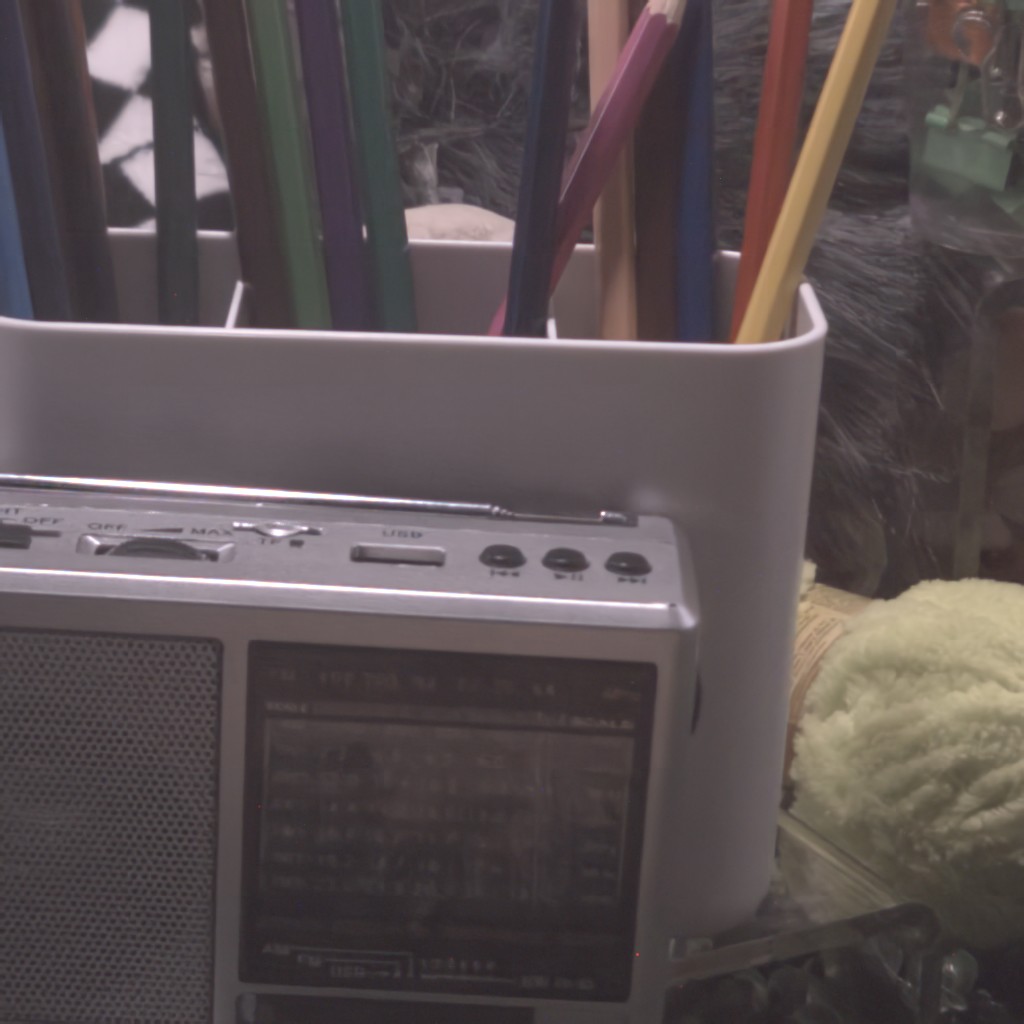}{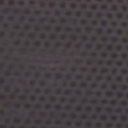}{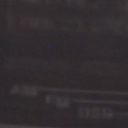} &
    \tile{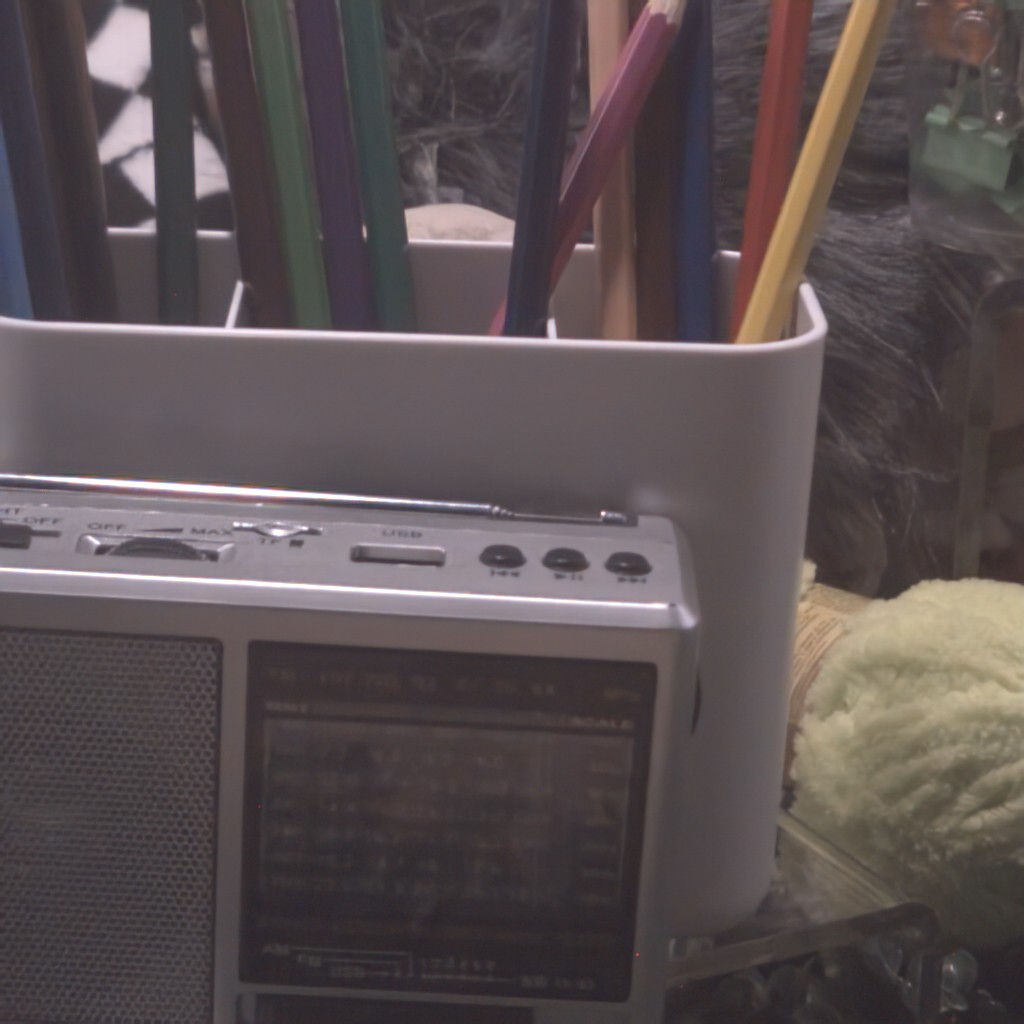}{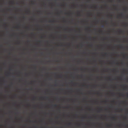}{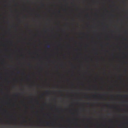} &
    \tile{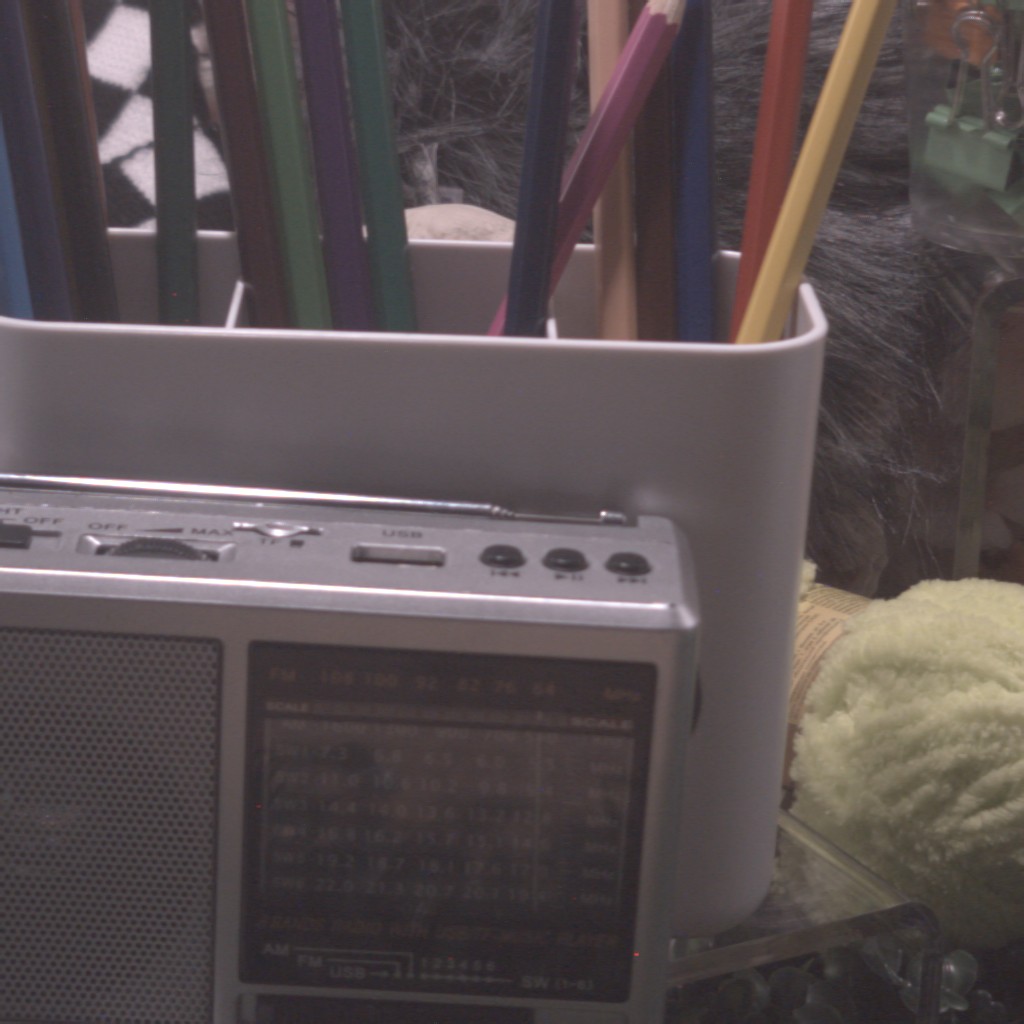}{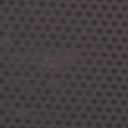}{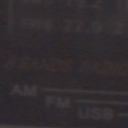} \\
    \vspace{-6pt}\\
    Noisy frame & VMCL-ISP & AxeraAI & XJAI & SNU-ISPL & Ground Truth
  \end{tabular}

  \caption{\textbf{Qualitative comparison.} Two sequences from final test data are selected for visual comparison. Viewer is advised to zoom in.}
  \label{fig:qualitative}
\end{figure*}

\section{Challenge Methods and Teams}
\subsection{SNU-ISPL}

The SNU-ISPL team proposes a low-light RAW video denoising framework, DarkVRAI, that exploits the prior information related to a given low-light noisy video frame sequence in both aligning and denoising process, and leverages the temporal relationship between frames with a new selective scan mechanism.

The prior information in the scope of this challenge are the sensor type, scene illuminance (lx), and frame rate (fps), but are not limited to these when additional information is available. These auxiliary cues provide critical information for both the alignment and denoising of low-light RAW video frames. Further referred as the `capture conditions'. 

Furthermore, they introduce a Burst-Order Selective Scan (BOSS) module that scans the frames in burst order to capture the long-range dependency of temporal information.

\begin{figure}[t]
    \centering
    \includegraphics[width=0.9\columnwidth]{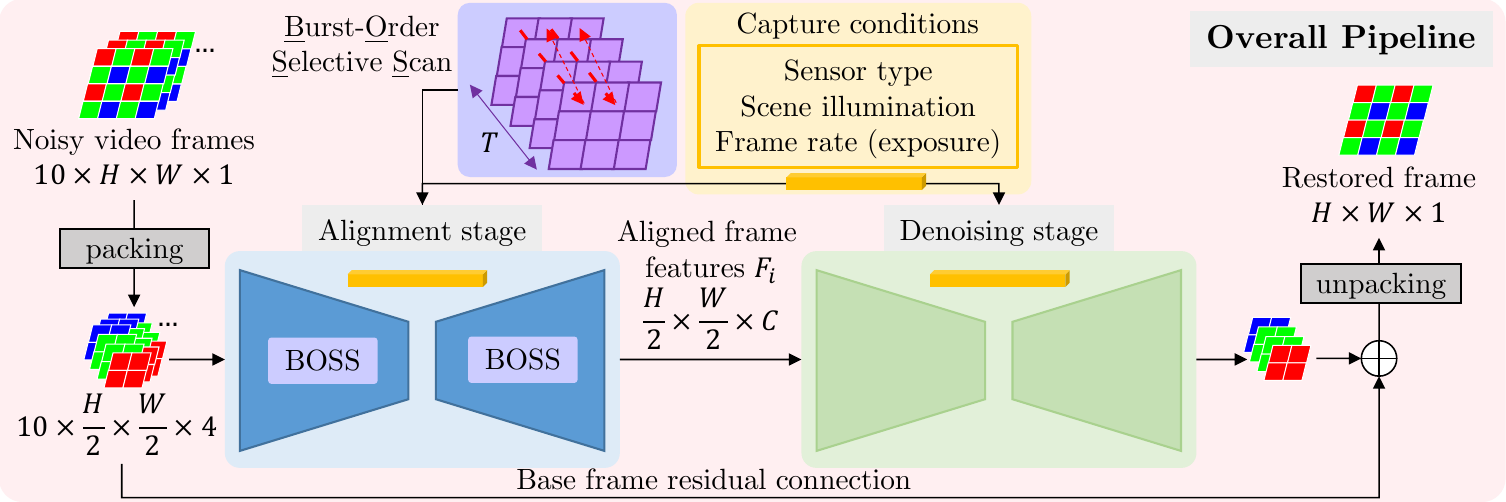}
    \caption{Overall pipeline of Team SNU-ISPL's proposed framework DarkVRAI for low-light RAW video denoising.}
    \label{fig:darkvrai_pipeline}
\end{figure}

\begin{figure}[t]
    \centering
    \includegraphics[width=0.9\columnwidth]{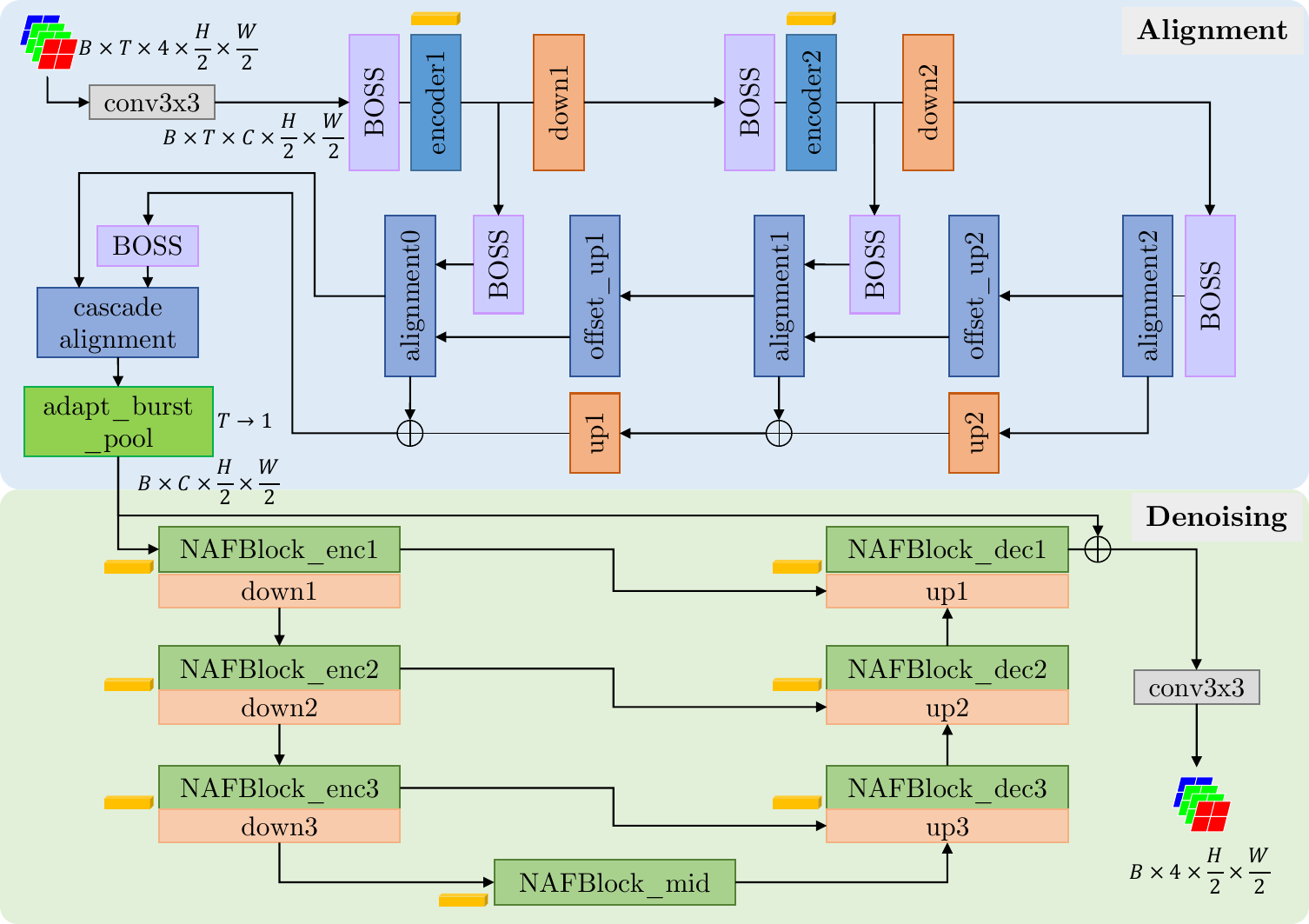}
    \caption{Team SNU-ISPL DarkVRAI network details.}
    \label{fig:darkvrai_network}
\end{figure}

Their framework DarkVRAI consists of two stages, each assisted by the capture conditions. The framework takes 10 noisy consecutive RAW frames as an input, and produces a single denoised RAW frame. 
The overall pipeline is shown in Fig.~\ref{fig:darkvrai_pipeline}, and each stage is illustrated in Fig.~\ref{fig:darkvrai_network}.

\noindent\textbf{Frame Alignment.} For the frame alignment, they build upon the alignment module in~\cite{dudhane2023burstormer}. The module is modified to produce aligned features $F_i$, with the number of channels $C$ as 48, from a sequence of 10 input RAW frames, where each frame is packed according to their Bayer pattern. BOSS module, which is based on~\cite{guo2024mambair}, is applied in front of each encoder and alignment stage to progressively aggregate long-range temporal cues, ensuring robust and accurate frame correspondence along the temporal axis.

\noindent\textbf{Denoising.} They choose convolutional NAFBlocks~\cite{chen2022simple} to build their network for the denoising stage. In this stage, the aligned and enriched features $F_i$ from the previous stage are further processed to remove noise and finally unpacked to produce a single restored RAW frame. The number of blocks for the encoder and decoder stages are 4,4,4 each, and 8 for the bottleneck stage.

\noindent\textbf{Conditioning with Capture Conditions.} Inspired by~\cite{oh2025towards}, each capture condition is first one-hot encoded and transformed into a trainable representative degradation vector, optimized through end-to-end training. This vector is then used to condition both alignment and denoising networks via adaptive layer normalization~\cite{peebles2023scalable}.

\noindent\textbf{Implementation Details.} The training data containing 5 scenes provided by the challenge organizers is cropped into patches of size 256 with stride 192. The proposed method is trained solely on this challenge data without further augmentation to preserve the Bayer pattern. The model is trained using Adam optimizer~\cite{kingma2015adam} ($\beta_1=0.9, \beta_2=0.999$) with an initial learning rate of 2e-4, gradually decaying to 1e-6 using Cosine Annealing scheduling~\cite{loshchilov2017sgdr}, for 300K iterations with batch size of 4 using a combination of L1 and MS-SSIM loss~\cite{wang2003multiscale} in an end-to-end manner. 
All experiments were performed using PyTorch framework and NVIDIA RTX 3090 GPUs.

\subsection{XJAI}

\begin{figure}[h]
    \centering
    \includegraphics[width=0.9\columnwidth]{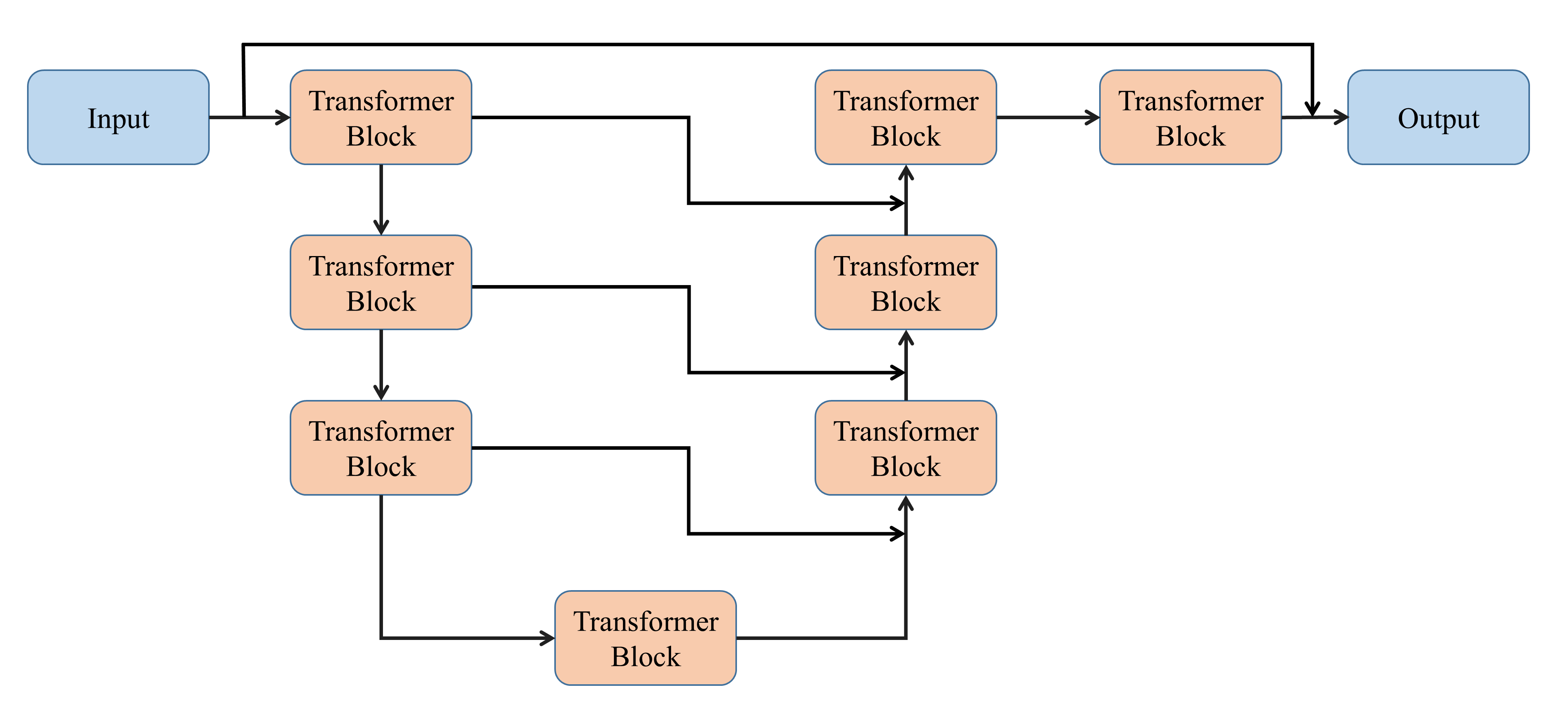}
    \caption{XJAI team pipeline schema.}
    \label{fig:XJAI}
\end{figure}

The XJAI team proposes a multi-level hierarchical framework based on efficient transformer architecture for high-resolution image restoration, which includes patch embedding, multi-scale feature extraction, and progressive refinement, see Figure~\ref{fig:XJAI}.

\noindent\textbf{Patch Embedding.} They begin with an overlapped patch embedding module that converts the input image into feature representations. Unlike traditional non-overlapping patch partitioning, their approach uses 3$\times$3 convolutions to generate overlapped patches, that better preserves local spatial information and reduces artifacts at patch boundaries.

\noindent\textbf{Multi-scale Feature Extraction.} Their framework employs a U-Net-like encoder-decoder architecture with four hierarchical levels. At each level, they utilize TransformerBlock modules adapted from Restormer~\cite{zamir2022restormer}. The encoder progressively downsamples features using PixelUnshuffle operations, allowing the model to capture both fine-grained local details and global contextual information. The number of transformer blocks at each level follows the configuration $[4,6,6,8]$, with attention heads scaling as $[1,2,4,8]$ to match the feature complexity.

\noindent\textbf{Progressive Refinement.} The decoder reconstructs high-resolution features through symmetric upsampling operations using PixelShuffle, with skip connections from corresponding encoder levels to preserve spatial details. Channel reduction convolutions are applied to manage feature concatenation efficiently. Finally, a refinement stage with 4 additional transformer blocks further enhances the restored image quality. The output layer includes a residual connection with the input image to facilitate stable training and better convergence.

\noindent\textbf{Training Strategy.} The training approach processes sequences of 9 consecutive frames as input with a random noise replacement strategy for enhanced robustness. For each training sample, they randomly select one of 20 additional noisy variants (\texttt{extra\_noisy\_09\_00} to \texttt{extra\_noisy\_09\_19}) with equal probability to replace the target frame, ensuring diverse noise exposure during training. They train the model using AdamW optimizer with learning rate 5$\times $$10^{-4}$, Charbonnier loss, and true cosine annealing schedule over 80\,000 iterations. The Restormer architecture is configured with 10 input channels for multi-frame processing and single-channel output, maintaining hierarchical transformer blocks [4,6,6,8] with bias-free LayerNorm for efficient high-resolution denoising.

\subsection{AxeraAI}
Considering the computational constraints (encompassing power consumption and processing capabilities) inherent in edge devices in the real world, convolutional neural network (Conv-based) models remain critically important. Consequently, they sought to quantitatively evaluate the achievable performance limits of Conv-based models. Following an exhaustive search across all potential candidates, they identified NAFNet \cite{chen2022simple} as exhibiting superior performance and consequently selected it as the baseline model for the experiments. To maximize performance, they implemented and evaluated three distinct methodological approaches:
\begin{itemize}

\item Supervised Single-Frame Training: They trained a NAFNet model from scratch using a purely supervised, single-frame approach, employing only the fundamental L1 loss function.

\item Self-Supervised Single-Frame Training: This approach utilized a purely self-supervised, single-frame methodology, drawing inspiration from the principles of Noise2Noise \cite{lehtinen2018noise2noise}.

\item Multi-Frame Finetuning: The single-frame model from Approach 1 served as a pre-trained foundation for the progressive training of a multi-frame model. Despite experimentation with various multi-frame fusion strategies, none yielded performance improvements over the baseline single-frame model. 

\end{itemize}
It is noted that exclusively the training data provided by the competition organizers was utilized throughout these experiments. During the validation phase, Approach 1 (supervised single-frame training) consistently outperformed the alternatives. Therefore, the results obtained using this method were directly utilized for the final testing phase evaluation. 

They train the model with a single NVIDIA 4090 GPU, which cost about 10 hours. They use Adam optimizer with a learning rate 1e-4. The batch size is 8, and the patch size is 256$\times$256$\times$4. The total epoch number is 2000, and they sample fixed 80 instances in every epoch.

\begin{figure}[h]
    \centering
    \includegraphics[width=1.\columnwidth]{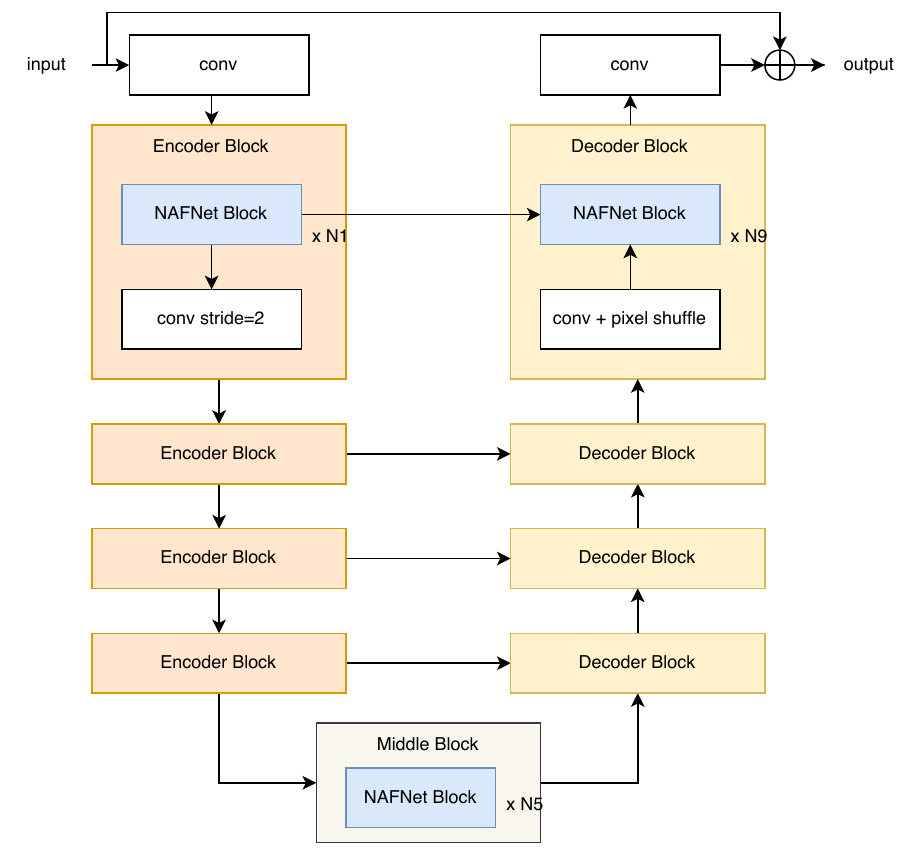}
    \caption{\textbf{Team AxeraAI pipeline.} They reuse the NAFNet constructed by the NAFNet Block.}
    \label{fig:AxeraAI}
\end{figure}

The whole pipeline can be found in Fig.~\ref{fig:AxeraAI}.


\subsection{VMCL-ISP}

The VMCL-ISP team's solution builds upon their previous work ``YOND: Practical Blind Raw Image Denoising Free from Camera-Specific Data Dependency"~\cite{yond}. They focus on data-centric approaches for raw image denoising, including dataset collection and noise modeling. In this section they first briefly outline proposed framework, then discuss their key data observations.

\noindent\textbf{Framework.} YOND follows traditional VST-based blind denoising pipeline: first utilizing coarse-to-fine noise estimation (CNE) to identify camera noise characteristics, then applying expectation-matched variance-stabilizing transform (EM-VST) to eliminate camera-specific data dependency, and finally employing an SNR-guided AWGN denoiser (SNR-Net) for the final denoising. To further improve noise estimation accuracy, they estimate noise parameters from color chart scenes in the training set under specific sensor and capture conditions, then apply corresponding noise parameters to textural test scenes. Further details can be found in \cite{yond}.

\begin{figure}[t]
	\centering
	\includegraphics[width=0.9\linewidth]{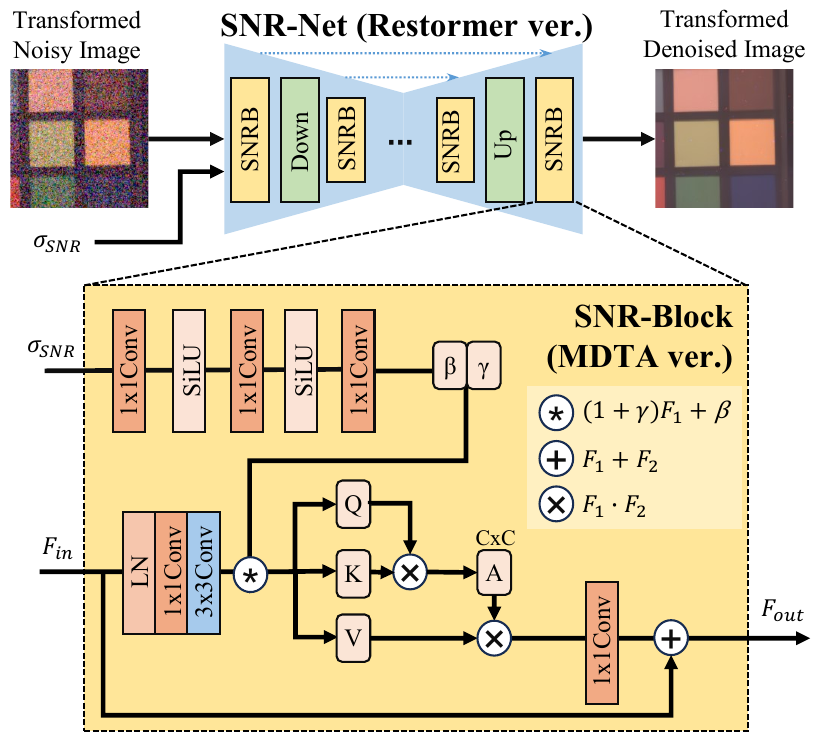}
	\caption{Detailed network structure of proposed SNR-Net.}
	\label{fig:SNR-Net}
\end{figure}

They adopt Restormer~\cite{zamir2022restormer} as the backbone for this SNR-Net. And augment Restormer with a guidance branch to fit the framework, as shown in Figure~\ref{fig:SNR-Net}. For training data, they adopt the LSDIR dataset~\cite{lsdir}, synthesizing pseudo-raw images via unprocessing and injecting clipped Gaussian noise.

\begin{figure}[t]
	\footnotesize
	\setlength\tabcolsep{2pt}
	\centering
	\begin{tabular}{cccc}
		Noisy Image & Denoised & Ground Truth & Denoised GT \\
		{\includegraphics[width=0.23\linewidth]{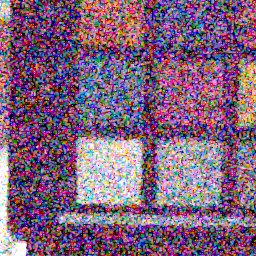}} &
		{\includegraphics[width=0.23\linewidth]{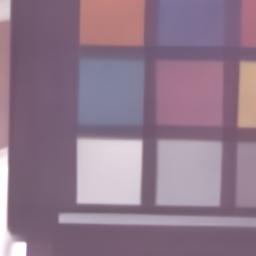}} &
		{\includegraphics[width=0.23\linewidth]{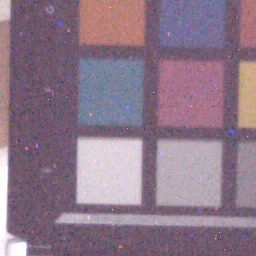}} &
		{\includegraphics[width=0.23\linewidth]{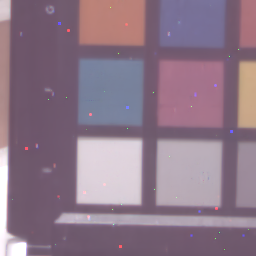}}
	\end{tabular}
	\caption{The visualization of data defects under low-light condition. ``Denoised" and ``Denoised GT" are blind denoised by the same YOND without additional training.}
	\label{fig:result}
	\vspace{-6pt}
\end{figure}

\noindent\textbf{Discussion.} As emphasized in their prior work PMN~\cite{pmn}, they advocate for improved data acquisition protocols to ensure data quality. This dataset employs multi-frame averaging (similar to SIDD~\cite{abdelhamed2018high}) to obtain the GT, which demonstrates limited effectiveness in low-light scenarios. As shown in Figure~\ref{fig:result}, they can identify a series of data defects from the comparison between the denoising results and the GT, including significant residual noise and defective pixels. Furthermore, some sensors exhibit systemic color bias due to black level errors, which means signals that should be zero are not normalized to zero. Calibrating the actual black level from noisy images (preferably dark frames) serves as a conventional solution. Unfortunately, the provided data has clipped negative noise values, which substantially complicates black level estimation. Therefore, they reiterate their call for heightened attention to the data quality in denoising.

\section{Conclusion}
We introduced the first AIM~2025 Low-Light RAW Video Denoising Challenge and a multi-device RAW video benchmark captured under nine illumination-exposure conditions with high-SNR references. Participants denoised linear RAW sequences and were ranked on a private test set using PSNR and SSIM. We expect the released dataset, protocol, and baselines to enable reproducible research and encourage further progress on RAW video denoising.

\section*{Acknowledgments}
This work was partially supported by the Alexander von Humboldt Foundation. We thank the AIM 2025 sponsors: AI Witchlabs and University of W\"urzburg (Computer Vision Lab).
The evaluation was carried out using the MSU-270 supercomputer of the Lomonosov Moscow State University.

{
    \small
    \bibliographystyle{ieeenat_fullname}
    \bibliography{main}
}

\clearpage
\appendix
\section{Teams and Affiliations}
\label{sec:teams}

\noindent \textit{\textbf{Team:}}\\
Organizers of AIM 2025 Low-light RAW Video Denoising Challenge\\
\textit{\textbf{Members:}}\\
Alexander Yakovenko$^{1,2}$ \\(\href{mailto:alexander.yakovenko@graphics.cs.msu.ru}{alexander.yakovenko@graphics.cs.msu.ru}), \\
George Chakvetadze$^1$ \\(\href{mailto:george.chakvetadze@graphics.cs.msu.ru}{george.chakvetadze@graphics.cs.msu.ru}), \\
Ilya Khrapov$^1$ (\href{mailto:ilya.khrapov@graphics.cs.msu.ru}{ilya.khrapov@graphics.cs.msu.ru}),\\
Maksim Zhelezov$^1$ \\(\href{mailto:maksim.zhelezov@graphics.cs.msu.ru}{maksim.zhelezov@graphics.cs.msu.ru}),\\
Dmitry Vatolin$^{1,2}$ (\href{mailto:dmitriy@graphics.cs.msu.ru}{dmitriy@graphics.cs.msu.ru}),\\
Radu Timofte$^3$ (\href{mailto:radu.timofte@uni-wuerzburg.de}{radu.timofte@uni-wuerzburg.de})\\ 
\textit{\textbf{Affiliations:}}\\
$^1$: Lomonosov Moscow State University, Russia\\
$^2$: MSU Institute for Artificial Intelligence, Russia \\
$^3$: University of Würzburg, Germany \\

\noindent \textit{\textbf{Team:}}\\
SNU-ISPL\\
\textit{\textbf{Members:}}\\
Youngjin Oh (\href{mailto:yjymoh0211@snu.ac.kr}{yjymoh0211@snu.ac.kr}),\\ 
Junhyeong Kwon (\href{mailto:gjh8760@snu.ac.kr}{gjh8760@snu.ac.kr}), \\
Junyoung Park (\href{mailto:parkjun21@snu.ac.kr}{parkjun21@snu.ac.kr}), \\
Nam Ik Cho (\href{mailto:nicho@snu.ac.kr}{nicho@snu.ac.kr})\\
\textit{\textbf{Affiliations:}}\\
Seoul National University, Republic of Korea\\

\newpage
\noindent \textit{\textbf{Team:}}\\
XJAI\\
\textit{\textbf{Members:}}\\
Senyan Xu (\href{mailto:syxu@mail.ustc.edu.cn}{syxu@mail.ustc.edu.cn}), \\
Ruixuan Jiang (\href{mailto:ruixuanjiang@mail.ustc.edu.cn}{ruixuanjiang@mail.ustc.edu.cn}), \\
Long Peng (\href{mailto:longp2001@mail.ustc.edu.cn}{longp2001@mail.ustc.edu.cn}), \\
Xueyang Fu (\href{mailto:xyfu@ustc.edu.cn}{xyfu@ustc.edu.cn}), \\
Zheng-Jun Zha (\href{mailto:zhazj@ustc.edu.cn}{zhazj@ustc.edu.cn})\\
\textit{\textbf{Affiliations:}}\\
University of Science and Technology of China, China\\

\noindent \textit{\textbf{Team: }}\\
AxeraAI\\
\textit{\textbf{Members:}}\\
Xiaoping Peng (\href{mailto:pengxiaoping@axera-tech.com}{pengxiaoping@axera-tech.com})\\
\textit{\textbf{Affiliations:}}\\
Axera Semiconductor Co., Ltd, China\\
\\
\noindent \textit{\textbf{Team:}}\\
VMCL-ISP\\
\textit{\textbf{Members:}}\\
Hansen Feng$^1$ (\href{mailto:fenghansen@bit.edu.cn}{fenghansen@bit.edu.cn}), \\
Zhanyi Tie$^1$ (\href{mailto:tiezhanyi@bit.edu.cn}{tiezhanyi@bit.edu.cn}), \\
Ziming Xia$^1$ (\href{mailto:xiazimin@bit.edu.cn}{xiazimin@bit.edu.cn}), \\
Lizhi Wang$^{2}$ (\href{mailto:wanglizhi@bnu.edu.cn}{wanglizhi@bnu.edu.cn})\\
\textit{\textbf{Affiliations:}}\\
$^1$: Beijing Institute of Technology, China\\
$^2$: Beijing Normal University, China \\

\end{document}